\documentclass{article}

\usepackage{microtype}
\usepackage{graphicx}
\usepackage{subfigure}
\usepackage{booktabs} % for professional tables
\usepackage{hyperref}
\usepackage{multirow}
\usepackage[para]{footmisc}
\usepackage{array}
\usepackage[table]{xcolor}
\usepackage{siunitx}
\usepackage{subfigure}

\usepackage{xspace}
\newcommand{\lmk}{\texttt{LMK}\xspace}
\usepackage[accepted]{icml2025}
\makeatletter
\renewcommand{\ICML@appearing}{}
\makeatother

\usepackage{amsmath}
\usepackage{amssymb}
\usepackage{mathtools}
\usepackage{amsthm}

\usepackage{algorithmic}
\usepackage{algorithm}
\usepackage{bbm}

\usepackage[capitalize,noabbrev]{cleveref}
\AddToHook{cmd/appendix/before}{\crefalias{section}{appendix}}

\theoremstyle{plain}

\theoremstyle{definition}

\theoremstyle{remark}

\usepackage[textsize=tiny]{todonotes}

\icmltitlerunning{Landmark Pooling for Dense Embeddings}

\begin{document}

\twocolumn[
\icmltitle{LMK $>$ CLS \\
           Landmark Pooling for Dense Embeddings}




\begin{icmlauthorlist}
\icmlauthor{Meet Doshi}{comp}
\icmlauthor{Aashka Trivedi}{comp}
\icmlauthor{Vishwajeet Kumar}{comp}
\icmlauthor{Parul Awasthy}{comp}
\icmlauthor{Yulong Li}{comp}
\break \centering
\icmlauthor{Jaydeep Sen}{comp}
\icmlauthor{Radu Florian}{comp}
\icmlauthor{Sachindra Joshi}{comp}
\end{icmlauthorlist}

\icmlaffiliation{comp}{IBM Research}

\icmlcorrespondingauthor{Meet Doshi}{meet@ibm.com}

\icmlkeywords{Dense Retrievers, Pooling Mechainsm, LMK Pooling, Long Context Embedding}

\vskip 0.3in
]



\printAffiliationsAndNotice{}  % leave blank if no need to mention equal contribution

\begin{abstract}
% Representation learning has been a key area in many important machine learning applications and serves as a key to improving performance in many applications like search, clustering, classification, reranking, etc. Most state-of-the-art methods work by using some pooling mechanism which converts a sequence of token embeddings into a single representative embedding. Most common approach is to either use a special token (preferably \texttt{[CLS]}) or use a mean pooling of all the token embeddings. In this work we highlight the pitfalls of existing pooling mechanisms and why they are biased for either short context or long context. To mitigate this, we propose Landmark (\texttt{[LMK]}) pooling which divides a text sequence into chunks and adds special tokens between them, the final representation is a mean pooled representation of all the \texttt{[LMK]} token representations. We show that this mechanism can perform as well as other pooling mechanism on short context retrieval tasks, whereas show a large performance increase in long context tasks. We posit that this pooling mechanism is a better alternative to existing methods.
Representation learning is central to many downstream tasks such as search, clustering, classification, and reranking. State-of-the-art sequence encoders typically collapse a variable-length token sequence to a single vector using a pooling operator—most commonly a special $\texttt{[CLS]}$ token or mean pooling over token embeddings. In this paper, we identify systematic weaknesses of these pooling strategies: $\texttt{[CLS]}$ tends to concentrate information toward the initial positions of the sequence and can under-represent distributed evidence, while mean pooling can dilute salient local signals, sometimes leading to worse short-context performance. To address these issues, we introduce \emph{Landmark} (\lmk) pooling, which partitions a sequence into chunks, inserts landmark tokens between chunks, and forms the final representation by mean-pooling the landmark token embeddings. This simple mechanism improves long-context extrapolation without sacrificing local salient features, at the cost of introducing a small number of special tokens. We empirically demonstrate that \lmk pooling matches existing methods on short-context retrieval tasks and yields substantial improvements on long-context tasks, making it a practical and scalable alternative to existing pooling methods.
\end{abstract}

\section{Introduction}
\label{introduction}

\begin{figure}[ht]
  \centering
  \includegraphics[width=0.85\linewidth]{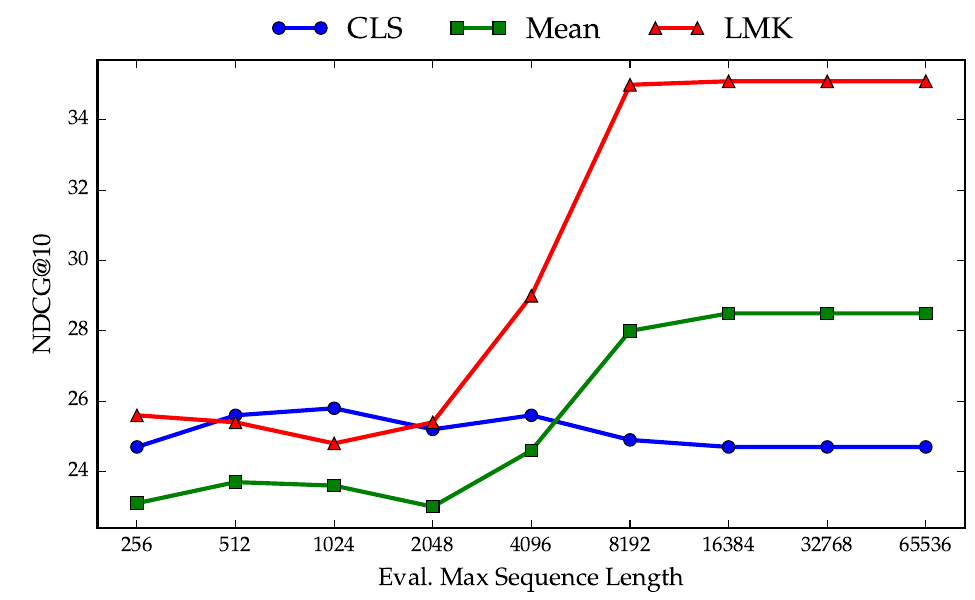}
\caption{MLDR (English) dataset long context retrieval performance for different pooling strategies using \emph{modernbert-base} finetuned on MSMarco Passages.}
  \label{fig:long_ctx_mldr}
\end{figure}

% \todo[inline]{Add missing citations and rewrite if necessary. Consistent use of LMK or \lmk, same for CLS}

 Traditionally, document-level representations were constructed using vector space based feature extraction methods such as CBOW and Skip gram, which aggregate word-level statistics or embeddings into fixed size representations \cite{skipgram_paper,clinchant-perronnin-2013-aggregating,word_embedding_ir}. These approaches were later supplanted by neural sequence encoders, including LSTMs and Transformers \cite{attention_paper,peters-etal-2018-deep}, primarily because classical methods fail to preserve rich document-level semantics and rely on fixed feature heuristics that cannot compete with contextualized representations learned by neural models. In particular, the Transformer architecture has consistently demonstrated superior performance over other neural sequence encoders in terms of representation quality and contextual understanding \citep{li-etal-2020-sentence}. Moreover, with successive architectural refinements, Transformers have been shown to be universal approximators of arbitrary sequence-to-sequence functions over compact domains \citep{transformers_universal_approx}.

A common application of neural text encoders is the construction of fixed-dimensional, sentence-level representations for downstream tasks such as retrieval, classification, and clustering \citep{li-etal-2020-sentence,yan-etal-2021-consert,labse_paper}. Formally, given an input text sequence $\mathcal{X}$, a tokenization function $\mathcal{F}_{\text{tok}}$ is first applied to obtain a sequence of tokens
$
\mathcal{X}_{\text{tok}} = (x_1, x_2, \ldots, x_t),
$
which is augmented with special markers such as \texttt{[CLS]} and \texttt{[SEP]} to denote the beginning and end of the sequence, respectively. The resulting token sequence is then passed through an encoder $\theta_{\text{enc}}$, producing a sequence of contextualized hidden representations
$
\mathbf{H}^{\text{enc}}\!=\!(\mathbf{h}_{\texttt{[CLS]}}, \mathbf{h}_1, \ldots, \mathbf{h}_t, \mathbf{h}_{\texttt{[SEP]}}),
$
where each vector corresponds to a token in the input sequence.

A key challenge is to transform this variable-length sequence of token representations $\mathbf{H}^{\text{enc}}$ into a single fixed-dimensional representation of the entire text sequence, denoted $\mathcal{X}^{\text{enc}}$. This is typically accomplished using a pooling function $\mathcal{P}$. Existing work most commonly adopts simple pooling strategies such as \texttt{[CLS]} pooling,
\begin{equation}
\label{eq:cls-pooling}
   \mathcal{P}_{\textsc{CLS}}(\mathbf{H}^{\text{enc}}) = \mathbf{H}^{\text{enc}}[0, :] = \mathbf{h}_{\texttt{[CLS]}} 
\end{equation}
or mean pooling,
\begin{equation}
\label{eq:mean-pooling}
    \mathcal{P}_{\textsc{mean}}(\mathbf{H}^{\text{enc}}) = \frac{1}{|\mathbf{H}^{\text{enc}}|} \sum_{i=0}^{|\mathbf{H}^{\text{enc}}|} \mathbf{h}_i.
\end{equation}

Intuitively, mean pooling provides a holistic aggregation by uniformly normalizing contributions from all token representations; however, this averaging can lead to a \emph{dilution} of salient or extreme features. In contrast, \texttt{[CLS]} pooling relies on a single special token to aggregate and encode all relevant information into a fixed-dimensional vector. While conceptually appealing, this approach places a heavy representational burden on a single position and, as we observe, often \emph{fails to generalize effectively to long contexts}, a property that is increasingly critical for modern text encoders.

To address these limitations, we investigate a simple yet effective alternative termed \emph{Landmark} (\lmk) pooling. Our approach is inspired by the method proposed by \citet{luo-etal-2024-landmark}, which introduces landmark tokens to mitigate abrupt chunking effects in retrieval-augmented generation (RAG) systems. In contrast, we extend this idea to the setting of Dense Passage Retrieval (DPR). Specifically, instead of relying on a single special token at the beginning of the sequence (e.g., \texttt{[CLS]}), we insert multiple special \emph{landmark} tokens at regular intervals after fixed-size chunks of input tokens. The final sequence representation is then obtained by applying mean pooling exclusively over the embeddings of these landmark tokens, rather than over all token embeddings. We posit that this strategy alleviates the representational bottleneck imposed by \texttt{[CLS]} in long-context settings, while simultaneously preserving salient local features that are often diluted under standard mean pooling.
We provide empirical evidence supporting these claims and demonstrate that \lmk pooling offers a substantially simpler yet effective mechanism for improving long-context generalization in neural text encoders, making it a practical alternative to commonly used pooling strategies.

\textbf{Our contributions:} 
% \textbf{(1)} We introduce \lmk pooling, a simple alternative to common pooling methods that yields substantial gains on long-context embedding benchmarks. 
% \textbf{(2)} Through extensive training and evaluation, we identify inherent limitations of popular pooling mechanisms for long-context generalization and characterize their strengths and weaknesses. 
% \textbf{(3)} We demonstrate that \lmk pooling is robust across training regimes and compatible with existing pretraining and retrieval optimizations, which can further improve its performance without degrading results on standard embedding tasks.
\textbf{(1)} We introduce \lmk pooling, a simple alternative to common pooling methods that delivers substantial gains on long-context embedding benchmarks across domains and languages.
\textbf{(2)} Through systematic evaluation, we identify inherent limitations of popular pooling mechanisms for long-context generalization and characterize their strengths and weaknesses. 
\textbf{(3)} We demonstrate that \lmk pooling is robust across training regimes and compatible with existing pretraining and retrieval optimizations.

\section{Related Work}
\label{related_work}

\textbf{Dense Embedders:} Early work on dense embedding representations \cite{early_dense_embedders_2,early_dense_embeddings} learned transformations of token representations by minimizing a distance objective between relevant queries and documents. These approaches were later superseded by stronger encoder architectures \cite{devlin-etal-2019-bert,modernbert} and improved training objectives \cite{infonce-paper}, which significantly enhanced representation quality. However, despite these advances, most methods still reduce a sequence of token representations to a single vector using simple pooling strategies, an aspect that has received little attention for long-context retrieval.

\textbf{Pooling Methodologies:} There is limited consensus in modern dense passage retrieval on how to aggregate token representations into a single document-level embedding. Many recent models rely on CLS pooling \cite{awasthy2025graniteembeddingr2models,zhang2024mgte,snowflake_arctic}, while others achieve comparable performance with mean pooling \cite{e5_paper,nomicembed}. Some adopt learned weighting or hybrid strategies \cite{bge-m3,nvembedv2}, with \citet{bge-m3} introducing multi-CLS pooling by inserting multiple CLS tokens at inference. Since these representations are fixed in size regardless of document length, they limit the amount of information that can be preserved \cite{colbertv2,limitations_dense_embedding}, motivating more robust pooling strategies. For long-context retrieval and retrieval-augmented generation, \citet{luo-etal-2024-landmark} introduce Landmark tokens to capture chunk-level representations in autoregressive language models; however, they do not investigate their application for pooling long documents into enhanced bidirectional dense embeddings. In contrast, our work systematically analyzes pooling mechanisms and evaluates their effectiveness across short- and long-context retrieval benchmarks. Additional studies have also highlighted limitations of existing pooling strategies; we refer readers to \cite{li-etal-2020-sentence,chen-etal-2023-ditto,nvembedv2,poolmewisely} for further discussion.

\textbf{Training and Evaluation Datasets:} Embedding training data have evolved from standard benchmarks such as MSMarco and NQ \cite{msmarco_dataset,kwiatkowski-etal-2019-natural} to large, heterogeneous mixtures used to train modern text embedders \cite{bge-m3,bge-en-icl-paper}. At the same time, evaluation has expanded to broad benchmarks spanning multiple domains, tasks, and languages \cite{beir-benchmark,mmteb}. However, most widely used benchmarks focus on relatively short sequences (typically $\leq$512 tokens). In contrast, many real world applications require robust evaluation on long context embedding tasks, which remain underexplored, particularly with respect to how representation and pooling choices affect performance \cite{coir-benchmark,zhu-etal-2024-longembed,chalkidis-etal-2021-multieurlex}. In this work, we explicitly connect pooling mechanisms with long context evaluation to systematically compare existing approaches against \lmk pooling.

% At the same time, evaluation has expanded from corresponding development and test sets \cite{trec_dl_paper} to broad benchmarks spanning multiple domains, tasks, and languages \cite{beir-benchmark,mmteb}

% information theoretic measures to evaluate embeddings \cite{layer_by_layer_icml25,} NOT NEEDED

\section{Understanding Pooling Methodologies}
\label{existing_pooling}

In this section, we analyze existing pooling methodologies as we motivate the need for \lmk pooling.

\subsection{CLS pooling}

% \todo[inline]{Begin with self-attention and rope embeddings, highlight how position is a critical part of generating embeddings— causing bias towards initial positions due to long term decay of RoPE embeddings, show empirical proofs on long context retrieval datasets like in the above fig where CLS gives more weightage to initial tokens. Also highlight and make connection with NTK scaling and why increasing this helps improve long context extrapolation performance for RoPE.}

\texttt{[CLS]} pooling is a widely used strategy in text embedding models. Most state-of-the-art approaches pretrain models on long-context masked language modeling (MLM) objectives and then fine-tune them for embedding tasks using \texttt{[CLS]} pooling, where most training datasets contain relatively short sequences. We show that \texttt{[CLS]} pooling inherently biases the model toward the early positions in the text, and that training on shorter sequences amplifies this bias. To understand the origin of this behavior, we analyze how positional information is encoded in text encoders.

In Transformers, to generate contextualized sequence representations, each input passes through an attention block. Let the input sequence be $\mathcal{X} = [x_1, \dots, x_t]$, with $x_i \in \mathbb{R}^d$. For a single attention head, we compute query, key, and value vectors:
\[
q_i = W_Q x_i, \quad k_i = W_K x_i, \quad v_i = W_V x_i,
\]
where $W_Q, W_K, W_V \in \mathbb{R}^{d \times d_h}$ and $d_h$ is head dimension. Attention scores, obtained from queries and keys, are used to weight the values to form contextualized representations.

Rotary Positional Embeddings (RoPE) \cite{rope_paper} encode positional information by applying a position-dependent rotation $R_{\theta,i}$ to the query and key vectors. The angular frequencies are defined as $\theta_j = b^{-\frac{2j}{d_h}}, \text{ for } j = 0, \dots, \frac{d_h}{2} - 1$, with $b$ as the base frequency (typically set to 10,000). Defining $\tilde{q}_m = R_{\theta,m} q_m$ and $\tilde{k}_n = R_{\theta,n} k_n$, the attention logit between positions $m$ and $n$ is:
\[
Attn(m,n)\!\coloneqq \tilde{q}_m^\top \tilde{k}_n = q_m^\top R_{\theta,m}^\top R_{\theta,n} k_n = q_m^\top R_{\theta,n-m} k_n,
\]
which shows that RoPE induces an explicit dependence on relative position, $n-m$.

\textbf{Long-term decay for CLS:} The rotation matrices in RoPE induce a relative distance-based attenuation of the attention logit $\tilde{q}_m^\top \tilde{k}_n$ for large relative offsets $|n-m|$, also referred to as the \emph{long-term decay} of RoPE. Specifically, for a \texttt{[CLS]} token at position $m=0$:
\[
\tilde{q}_{\text{CLS}}^\top \tilde{k}_n = q_{\text{CLS}}^\top R_{\theta,n} k_n,
\]
whose magnitude decreases as $n$ increases. After the softmax operation, this results in smaller attention weights for distant tokens, biasing the \texttt{[CLS]} representation toward early positions. This effect becomes more pronounced with longer sequences, limiting the effectiveness of \texttt{[CLS]} pooling for long-context extrapolation. Increasing the RoPE base frequency ($b$) slows the positional decay and can improve long-context extrapolation at inference \cite{zhang2024mgte,yarn-paper}, however, this does not fully mitigate the underlying bias.

\begin{figure}[ht]
  \centering
  \includegraphics[width=1.0\linewidth]{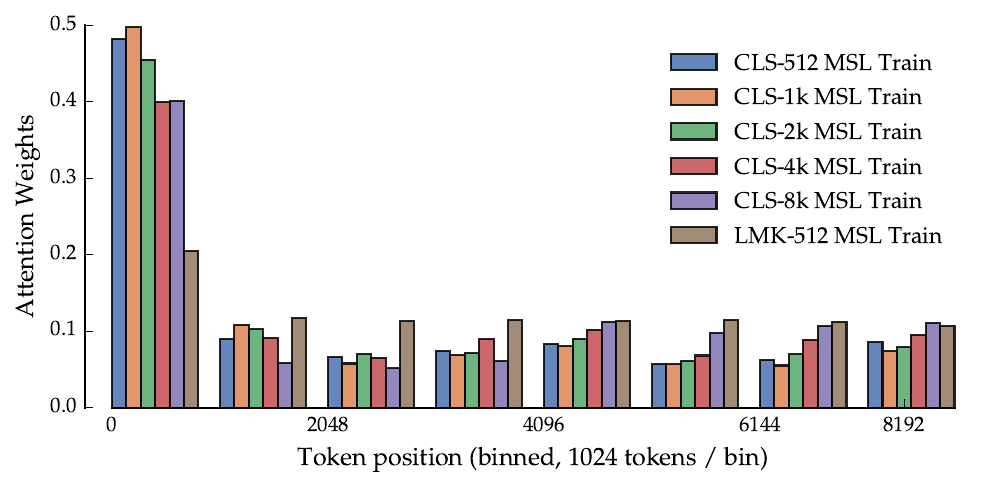}
  \caption{Normalized attention distribution over long MLDR documents ($>8192$ tokens) for \emph{mmBERT-base} fine-tuned with \texttt{CLS} and \lmk pooling with varying maximum sequence lengths (MSL).}
  \label{fig:cls_attn_span}
\end{figure}

We provide empirical evidence of this bias in \texttt{[CLS]} pooling in \cref{fig:cls_attn_span}. By analyzing the final-layer attention weights assigned to pooling tokens (\texttt{[CLS]} and \lmk) across long documents sampled from the MLDR English subset, we observe a distinct concentration of attention toward early tokens in the sequence. Fine-tuning on longer sequences reduces this bias but does not eliminate it. The effect worsens significantly when extrapolating beyond the training sequence length. In contrast, our \lmk pooling strategy shows no such bias despite being trained on only 512 tokens. Although these results are visualized for the final layer, the findings remain relevant for recent architectures like ModernBERT \cite{modernbert}, where attention to all tokens is restricted to specific layers.

\subsection{Mean and Latent Attention pooling}
\label{subsec:mean_latent_pooling}

Given a sequence of token embeddings $\mathbf{H}^{\text{enc}} \in \mathbb{R}^{S \times D}$, mean pooling (defined in \cref{eq:mean-pooling}) computes the average across all token embeddings in each latent dimension. Unlike \texttt{[CLS]} pooling, which relies on positional interactions, mean pooling is position-independent. While this simplicity is appealing, it encourages token representations to collapse toward a shared centroid in the embedding space, which can weaken salient features.  Moreover, mean pooling assigns equal weights to all tokens regardless of their semantic importance, thus tokens carrying little task-relevant information contribute uniformly to final embeddings in typical $\ell_2$-normalized (norm bounded) encoder representations.

% By assigning equal weight to all tokens regardless of their semantic importance, mean pooling treats informative and uninformative tokens identically. Since encoder representations are typically $\ell_2$-normalized (norm bounded), even tokens carrying little task-relevant information contribute uniformly to the final embedding. 

This uniform aggregation implicitly assumes that task relevant information is evenly distributed across the sequence, an assumption that often does not hold in practice, especially for short context tasks where critical evidence may be concentrated in a small subset of tokens. Consequently, mean pooling can underperform approaches such as \texttt{[CLS]} pooling, which allow the model to learn an explicit aggregation mechanism through attention. We observe that mean pooling extrapolates well to long context at inference time, and it faces no inherent architectural limitations in doing so; however, it consistently underperforms on short context tasks relative to \texttt{[CLS]} pooling. 

\citet{nvembedv2} introduce \emph{Latent Attention} to address the limitation of assigning equal importance to all tokens. This method uses latent vectors to determine how to attend selectively to different token embeddings in the sequence. Formally, let
$\mathbf{H}^{\text{enc}} = (\mathbf{h}_{\texttt{[CLS]}}, \mathbf{h}_1, \ldots, \mathbf{h}_t, \mathbf{h}_{\texttt{[SEP]}})$ denote the sequence of embeddings produced by the encoder, where each $\mathbf{h}_i \in \mathbb{R}^d$. Let $\mathbf{L} \in \mathbb{R}^{\ell \times d_l}$ denote a set of learnable latent vectors, where $\ell$ is the number of latents and $d_l$ is the latent head dimension. A latent attention module first computes the query, key, and value representations for $\mathbf{H}^{\text{enc}}$ and $\mathbf{L}$ as
\[
q_i = W_Q \mathbf{h}_i, \quad k = W_K \mathbf{L}, \quad v = W_V \mathbf{L},
\]
where $W_Q \in \mathbb{R}^{d \times d_h}$ and $W_K, W_V \in \mathbb{R}^{(\ell \times d_l) \times d_h}$, and $d_h$ denotes the head dimension corresponding to $\mathbf{h}_i$. 

\begin{equation}
\begin{gathered}
\mathcal{Y}_i = \text{Cross-Attn}\!\left(\text{Prenorm}(q_i), k, v\right) + q_i, \\
\mathcal{Z}_i = \text{FFN}\!\left(\text{Prenorm}(\mathcal{Y}_i)\right) + \mathcal{Y}_i, \\
\mathcal{P}_{\textsc{latent-attn}}
= \frac{1}{|\mathcal{Z}|} \sum_{i=0}^{|\mathcal{Z}|} \mathcal{Z}_i .
\end{gathered}
\end{equation}

Here, Prenorm refers to standard LayerNorm applied prior to the transformation, and FFN denotes the standard feed-forward network used in Transformer architectures.

This approach introduces additional parameters, including learnable latent vectors and feed-forward layers, which are used to determine how the latents are reweighted based on the sequence representations to produce new embeddings. The representations generated from these latent vectors are then mean pooled to obtain a single sequence-level representation. Conceptually, this procedure is equivalent to mean pooling and, as such, should not introduce a bias toward long-context extrapolation, which is consistent with our empirical results in \cref{results}. We therefore posit that latent attention inherits the same limitations as mean pooling, while incurring additional parameter and computational overhead that yields limited empirical gains.

\section{Landmark (LMK) Pooling}
\label{lmk_pooling}

\begin{algorithm}[ht]
\caption{Landmark Tokenization}
\label{alg:lmk_tokenize}
\begin{algorithmic}[1]
\small
\INPUT Text sequence $\mathcal{X}$, Tokenizer $\mathcal{F}_{\text{tok}}$, Granularity $g$, Landmark token $t_{\text{LMK}}$
\OUTPUT Token IDs $\mathcal{X}_{\text{tok}}$, Attention mask $\mathcal{X}_{\text{mask}}$
% \STATE $S \leftarrow \text{Split}(\mathcal{X}, g)$ \hfill \COMMENT{Split by sentence or fixed length}
% \STATE $Z \leftarrow \{ \mathcal{F}_{\text{tok}}(s) \mid s \in S \}$ \hfill \COMMENT{Tokenize (no special tokens)}
\STATE $T \leftarrow \mathcal{F}_{\text{tok}}(\mathcal{X})$ \hfill \COMMENT{Tokenize (no special tokens)}
\STATE $Z \leftarrow \text{Split}(T, g)$ \hfill \COMMENT{Split token sequence}
\STATE $\mathcal{X}_{\text{tok}} \leftarrow [t_{\text{CLS}}] \oplus z_1 \oplus [t_{\text{LMK}}] \oplus z_2 \dots \oplus [t_{\text{LMK}}] \oplus z_n \oplus [t_{\text{LMK}}]$
\STATE $\mathcal{X}_{\text{mask}} \leftarrow \mathbbm{1}(\mathcal{X}_{\text{tok}} \neq t_{\text{PAD}})$
\STATE \textbf{return} $\mathcal{X}_{\text{tok}}, \mathcal{X}_{\text{mask}}$
\end{algorithmic}
\end{algorithm}

\begin{algorithm}[ht]
\caption{Landmark (\lmk) Pooling}
\label{alg:lmk_pool}
\begin{algorithmic}[1]
\small
\INPUT Token embeddings $\mathbf{H}^{\text{enc}}\in\mathbb{R}^{S\times D}$, Token IDs $\mathcal{X}_{\text{tok}}$, Attention mask $\mathcal{X}_{\text{mask}}$, Landmark token $t_{\text{LMK}}$
\OUTPUT Sentence embedding $\mathcal{X}^{\text{enc}}\in\mathbb{R}^{D}$
\STATE $\mathcal{I}_L \leftarrow \{\, i \mid \mathcal{X}_{\text{tok}, i} = t_{\text{LMK}} \ \land\ \mathcal{X}_{\text{mask}, i}=1 \,\}$ \hfill \COMMENT{LMK index set}
\STATE $\mathcal{X}^{\text{enc}} \leftarrow \frac{1}{|\mathcal{I}_L|} \sum_{i \in \mathcal{I}_L} \mathbf{H}_{i,:}^{\text{enc}}$ \hfill \COMMENT{Mean over LMK}
\STATE \textbf{return} $\mathcal{X}^{\text{enc}}$
\end{algorithmic}
\end{algorithm}

% \todo[inline]{This section has a lot of repeated sentences that are similar to statements made in intro.}

% Having reviewed common pooling strategies, we highlight
% several key limitations shared by these approaches. First, \texttt{CLS} pooling exhibits an inherent bias toward tokens that are closer in relative position, as attention to distant tokens is more strongly attenuated. At the same time, the \texttt{CLS} token serves as a learned selector that enables the model to adaptively attend to different parts of the sequence, which is a desirable property for representation learning. Second, approaches that allow each token to attend to the sequence but then aggregate these representations via a simple expectation (e.g., mean pooling) tend to dilute salient features. Nevertheless, enabling multiple tokens to independently assess the importance of different parts of the sequence is another important characteristic of effective text embedders. Third, at inference time, we aim to avoid additional computational overhead or reliance on frozen latent vectors to reweigh sequence embeddings, as such mechanisms place disproportionate emphasis on latent features and increase complexity.

We highlight the key limitations of these pooling approaches, while noting the benefit that they provide. \texttt{CLS} pooling biases attention towards nearer tokens, attenuating the attention from distant positions. However, the \texttt{CLS} token acts as a learned selector that adaptively attends to different parts of the sequence- a desirable feature in representation learning. Furthermore, mean pooling dilutes salient features by uniformly averaging over all tokens, yet it allows multiple tokens to independently asses the full sequence, which may be more effective than single-token \texttt{CLS} representation. Finally, latent attention relies on frozen latent vectors to reweigh sequence embeddings, thereby adding undesirable complexity at inference time.

Landmark (\lmk) pooling provides a simple yet effective solution to the limitations discussed above. It introduces two key modifications to standard text encoding: (1) a change to the tokenization procedure and (2) a pooling strategy that depends on the positions of \lmk tokens. Instead of relying on a single special token, \lmk pooling inserts multiple special tokens throughout the input sequence. \textbf{We use the SEP token as the LMK token, as it already serves as a natural delimiter in most language encoders\footnote{For tokenizers that do not have a \texttt{SEP} token, the \texttt{EOS} token is an appropriate alternative.}}.

% We explore several strategies for placing these tokens along the sequence. One option is to split the text into sentences prior to tokenization and insert \lmk tokens after each sentence. However, this approach may not generalize well to other domains, such as code or certain natural languages, where clear sentence delimiters may be absent. Alternatively, \lmk tokens can be inserted at fixed intervals after tokenization, but this can introduce positional bias while making performance sensitive to the granularity chosen during fine-tuning. To address these issues, we also adopt a variable-granularity strategy during training, where the \lmk insertion interval is randomly sampled. This encourages robustness to granularity choice and allows a single model to support different granularities at inference depending on the task. The final sequence representation is obtained by mean pooling the embeddings of all (\lmk) tokens.

We explore several strategies for placing these tokens along the sequence. \textbf{Sentence Chunking} involves splitting the text into sentence prior to tokenization and inserting \lmk tokens after each sentence. However, this approach may not generalize well to other domains, such as code or certain natural languages, where clear sentence delimiters may be absent. Alternatively, \textbf{fixed length chunking} inserts \lmk tokens at fixed intervals after tokenization, but this can introduce positional bias while making performance sensitive to the granularity chosen during fine-tuning.
To address these issues, \textbf{variable chunking} randomly samples the \lmk insertion interval from a fixed set of predefined intervals. This encourages robustness to granularity choice and allows a single model to support different granularities at inference depending on the task. The final sequence representation is obtained by mean pooling the embeddings of all \lmk tokens.

While inserting additional special tokens increases the effective sequence length and incurs extra computation, the overhead is modest in practice since \lmk tokens are added at coarse intervals. For example, on the LongEmbed benchmark with a maximum evaluation sequence length (MSL) of 32K tokens, inserting an \lmk token every 128 tokens adds only 256 additional tokens to the sequence while significantly outperforming alternative pooling mechanisms as seen in \cref{tab:modernbert_en_lmk_pooling_results}. Formal pseudocode for the tokenization procedure is provided in \cref{alg:lmk_tokenize}, and the pooling operation is detailed in \cref{alg:lmk_pool}.

\section{Experimental Setup}
\label{exp_setup}

In this section, we describe the experimental setup used to obtain our results. First, we train models on standard retrieval datasets using different pooling strategies and compare their performance on widely used retrieval benchmarks. Second, we analyze the effect of long-context evaluation across different domains. Third, we study how long-context training influences the extrapolation behavior of pooling mechanisms. We divide our experiments into \textbf{English} and \textbf{Multilingual} settings, and further into short- and long-context training.

\paragraph{English:} 
We select widely used Transformer-based text encoders that support long-context training and fine-tuning, namely \emph{\href{https://huggingface.co/Alibaba-NLP/gte-en-mlm-base}{gte-en-mlm-base}} and \emph{\href{https://huggingface.co/answerdotai/ModernBERT-base}{ModernBERT-base}}. Standard retrieval training datasets, including MS MARCO passage and document ranking as provided by \citet{bge-en-icl-paper}, are used. Training includes hard negatives and distillation scores, with 7 hard negatives for passage ranking and 1 for document ranking. Models are trained for 5k steps with a learning rate of $2\times10^{-5}$. Passage ranking uses an effective query batch size of 2,048 with a maximum sequence length of 512 tokens, while document ranking uses a batch size of 256 with sequences up to 8,192 tokens.

To mitigate bias from domain overlap, we evaluate on a diverse set of out of domain benchmarks. In domain performance is reported on the MS MARCO Dev set, followed by short context evaluations on BEIR \cite{beir-benchmark}, MTEB-v2 \cite{mmteb}, and MIRACL Retrieval \cite{miracl-dataset}. For long context evaluation, we use MLDR, the COIR code retrieval benchmark \cite{coir-benchmark}, and LongEmbed, which includes documents of diverse lengths, often exceeding the 8k token pretraining context, with dispersed target information \cite{zhu-etal-2024-longembed}.

\paragraph{Multilingual:} 
For the multilingual setting, we use the multilingual counterpart of ModernBERT, \emph{\href{https://huggingface.co/jhu-clsp/mmBERT-base}{mmBERT-base}}. Models are fine-tuned on multilingual training data from \citet{bge-m3}, which includes diverse datasets with hard negatives and distillation scores generated by a multilingual re-ranker. Training is performed for 10k steps with a learning rate of $2\times10^{-5}$, using an effective query batch size of 1024, maximum sequence length of 512 tokens, and 7 hard negatives. In ablations, we investigate fine-tuning with sequence lengths up to 8{,}192 tokens using \emph{mmBERT-base} and \emph{\href{https://huggingface.co/Alibaba-NLP/gte-multilingual-base}{gte-multilingual-base}}, reducing the batch size to 64.

We evaluate multilingual retrieval on short-context benchmarks such as MIRACL Hard Negatives (18 languages) and long-context benchmarks like Multilingual Long Document Retrieval (MLDR, 13 languages), additionally reporting results on LongEmbed. Zero-shot multilingual long-document classification is evaluated on Multi-EURLEX across 23 languages \cite{chalkidis-etal-2021-multieurlex}, reporting Macro-F1 scores.

\paragraph{Pooling Strategies:} 
We compare \lmk pooling against multiple baselines from prior work, including \texttt{CLS}, Mean, and Latent Attention (Attn.) pooling with 64 latents, as described in \cref{subsec:mean_latent_pooling}. We also evaluate \textbf{MultiCLS}, where a model trained with \texttt{CLS} pooling is modified at inference by inserting \texttt{CLS} tokens at fixed intervals and mean-pooling their representations, analogous to \lmk at inference but without \lmk-style training. As shown by \citet{bge-m3}, this approach improves MLDR performance for \texttt{CLS} in the absence of long-context training. In addition, we include a \textbf{Mean@k} baseline, which computes the final representation by mean pooling every $k$-th token, to assess whether special pooling tokens are necessary for strong performance.

For Landmark pooling, we compare different insertion granularities of \lmk tokens, as described in \cref{lmk_pooling}. We consider a sentence-based splitter using a heuristic English sentence splitter\footnote{\href{https://github.com/mediacloud/sentence-splitter}{github.com/mediacloud/sentence-splitter}} to study the role of sentence boundaries, a fixed splitter that inserts \lmk tokens at regular intervals after tokenization, and a variable splitter where the granularity is uniformly sampled from $\{32, 64, 128, 256\}$ during training. This provides robustness and enables a single model to support multiple granularities at inference time. All pooling strategies are trained with identical hyperparameters and under the same data regime to eliminate training-related biases. During evaluation, we use a maximum sequence length of 8{,}192 for all tasks except LongEmbed, which uses 32{,}768, and explicitly report pooling strategies and granularities in the corresponding tables and figures. \textbf{Importantly, for LMK evaluations, we do not increase the token budget to account for additional special tokens}. We report standard retrieval metrics such as NDCG@10 and Precision@1. Additional training and evaluation details are provided in \cref{app:train_eval_details}, with dataset descriptions in \cref{app:dataset_details}.

\begin{table*}[ht]
\centering
\small
\setlength{\tabcolsep}{5pt}
\renewcommand{\arraystretch}{1.15}

\resizebox{\textwidth}{!}{%
\begin{tabular}{l c l c
                c
                c c c c
                c c c c}
\toprule

 & & & &
\multicolumn{1}{c}{\textsc{In-Domain}} &
\multicolumn{4}{c}{\textsc{Out-of-Domain (Short Context)}} &
\multicolumn{4}{c}{\textsc{Out-of-Domain (Long Context)}} \\

\cmidrule(lr){5-5}
\cmidrule(lr){6-9}
\cmidrule(lr){10-13}

& & & &
\textbf{MSMarco} &
\textbf{BEIR} &
\textbf{MTEBv2} &
\textbf{MIRACL} &
\textbf{Avg.} &
\textbf{MLDR} &
\textbf{COIR} &
\textbf{LongEmbed} &
\textbf{Avg.} \\

\multicolumn{2}{c}{\textsc{Finetuning}} &
\multicolumn{2}{c}{\textsc{Evaluation}} &
\textbf{Dev} &
\textbf{(15)} &
\textbf{(10)} &
\textbf{Dev} &
\textbf{Short} &
\textbf{Test} &
\textbf{(8)} &
\textbf{(6)} &
\textbf{Long} \\

\cmidrule(lr){1-2}
\cmidrule(lr){3-4}

\textbf{Pool} &
\textbf{Gran.} &
\textbf{Pool} &
\textbf{Gran.} &
\texttt{NDCG@10} &
\texttt{NDCG@10} &
\texttt{NDCG@10} &
\texttt{NDCG@10} &
 &
\texttt{NDCG@10} &
\texttt{NDCG@10} &
\texttt{P@1\;|\;NDCG@10} &
 \\

\midrule

CLS & - & CLS & - & \cellcolor[rgb]{0.749,0.833,0.998}{39.8} & \cellcolor[rgb]{0.616,0.712,0.972}{42.8} & \cellcolor[rgb]{0.642,0.739,0.984}{44.2} & \cellcolor[rgb]{0.461,0.509,0.828}{48.7} & \cellcolor[rgb]{0.558,0.644,0.934}{45.2} & \cellcolor[rgb]{0.857,0.894,0.953}{24.9} & \cellcolor[rgb]{0.824,0.879,0.974}{43.5} & \cellcolor[rgb]{0.794,0.311,0.405}{43.3} & \cellcolor[rgb]{0.977,0.812,0.737}{37.2} \\
CLS & - & MultiCLS & 128 & \cellcolor[rgb]{0.891,0.531,0.489}{39.0} & \cellcolor[rgb]{0.794,0.311,0.405}{33.6} & \cellcolor[rgb]{0.794,0.311,0.405}{31.2} & \cellcolor[rgb]{0.794,0.311,0.405}{31.0} & \cellcolor[rgb]{0.794,0.311,0.405}{31.9} & \cellcolor[rgb]{0.794,0.311,0.405}{5.1} & \cellcolor[rgb]{0.794,0.311,0.405}{35.4} & \cellcolor[rgb]{0.802,0.336,0.411}{43.9} & \cellcolor[rgb]{0.794,0.311,0.405}{28.1} \\
Mean & - & Mean & - & \cellcolor[rgb]{0.749,0.833,0.998}{39.8} & \cellcolor[rgb]{0.796,0.864,0.987}{41.6} & \cellcolor[rgb]{0.752,0.835,0.998}{43.3} & \cellcolor[rgb]{0.516,0.589,0.895}{48.0} & \cellcolor[rgb]{0.665,0.762,0.991}{44.3} & \cellcolor[rgb]{0.734,0.821,0.999}{28.0} & \cellcolor[rgb]{0.760,0.840,0.996}{44.1} & \cellcolor[rgb]{0.461,0.509,0.828}{62.7} & \cellcolor[rgb]{0.627,0.724,0.978}{44.9} \\
Latent Attn & - & Latent Attn & - & \cellcolor[rgb]{0.544,0.626,0.922}{40.1} & \cellcolor[rgb]{0.838,0.886,0.967}{41.3} & \cellcolor[rgb]{0.774,0.851,0.994}{43.1} & \cellcolor[rgb]{0.490,0.552,0.865}{48.3} & \cellcolor[rgb]{0.680,0.776,0.995}{44.2} & \cellcolor[rgb]{0.873,0.899,0.940}{24.5} & \cellcolor[rgb]{0.844,0.889,0.962}{43.3} & \cellcolor[rgb]{0.461,0.509,0.828}{62.8} & \cellcolor[rgb]{0.711,0.803,0.999}{43.5} \\
Mean@k & 64 & Mean@k & 64 & \cellcolor[rgb]{0.794,0.311,0.405}{38.6} & \cellcolor[rgb]{0.978,0.805,0.729}{39.7} & \cellcolor[rgb]{0.941,0.884,0.852}{41.4} & \cellcolor[rgb]{0.904,0.906,0.907}{44.1} & \cellcolor[rgb]{0.946,0.880,0.844}{41.7} & \cellcolor[rgb]{0.932,0.614,0.545}{15.7} & \cellcolor[rgb]{0.941,0.634,0.561}{39.8} & \cellcolor[rgb]{0.850,0.455,0.448}{44.8} & \cellcolor[rgb]{0.897,0.543,0.496}{33.4} \\
\midrule

\multirow{4}{*}{LMK} & 32 & \multirow{4}{*}{LMK} & 32 & \cellcolor[rgb]{0.817,0.876,0.978}{39.7} & \cellcolor[rgb]{0.464,0.514,0.832}{43.9} & \cellcolor[rgb]{0.461,0.509,0.828}{45.9} & \cellcolor[rgb]{0.470,0.523,0.840}{48.5} & \cellcolor[rgb]{0.461,0.509,0.828}{46.1} & \cellcolor[rgb]{0.461,0.509,0.828}{34.8} & \cellcolor[rgb]{0.510,0.580,0.888}{46.4} & \cellcolor[rgb]{0.576,0.666,0.947}{60.2} & \cellcolor[rgb]{0.506,0.575,0.884}{47.1} \\
 & 64 &  & 64 & \cellcolor[rgb]{0.680,0.776,0.995}{39.9} & \cellcolor[rgb]{0.516,0.589,0.895}{43.5} & \cellcolor[rgb]{0.537,0.617,0.916}{45.1} & \cellcolor[rgb]{0.500,0.566,0.876}{48.2} & \cellcolor[rgb]{0.513,0.585,0.892}{45.6} & \cellcolor[rgb]{0.506,0.575,0.884}{33.6} & \cellcolor[rgb]{0.490,0.552,0.865}{46.6} & \cellcolor[rgb]{0.583,0.674,0.952}{60.0} & \cellcolor[rgb]{0.527,0.603,0.905}{46.7} \\
 & 128 &  & 128 & \cellcolor[rgb]{0.609,0.704,0.968}{40.0} & \cellcolor[rgb]{0.601,0.695,0.964}{42.9} & \cellcolor[rgb]{0.665,0.762,0.991}{44.0} & \cellcolor[rgb]{0.480,0.538,0.853}{48.4} & \cellcolor[rgb]{0.569,0.657,0.942}{45.1} & \cellcolor[rgb]{0.616,0.712,0.972}{30.8} & \cellcolor[rgb]{0.520,0.594,0.898}{46.3} & \cellcolor[rgb]{0.873,0.899,0.940}{54.1} & \cellcolor[rgb]{0.699,0.793,0.998}{43.7} \\
 & 256 &  & 256 & \cellcolor[rgb]{0.680,0.776,0.995}{39.9} & \cellcolor[rgb]{0.767,0.845,0.995}{41.8} & \cellcolor[rgb]{0.810,0.872,0.981}{42.8} & \cellcolor[rgb]{0.616,0.712,0.972}{47.0} & \cellcolor[rgb]{0.715,0.806,1.000}{43.9} & \cellcolor[rgb]{0.486,0.547,0.861}{34.1} & \cellcolor[rgb]{0.510,0.580,0.888}{46.4} & \cellcolor[rgb]{0.673,0.769,0.993}{58.2} & \cellcolor[rgb]{0.554,0.639,0.931}{46.2} \\
\midrule

\multirow{4}{*}{LMK}
& \multirow{4}{*}{Variable}
& \multirow{4}{*}{LMK} & 32 & \cellcolor[rgb]{0.461,0.509,0.828}{40.3} & \cellcolor[rgb]{0.461,0.509,0.828}{44.0} & \cellcolor[rgb]{0.464,0.514,0.832}{45.8} & \cellcolor[rgb]{0.461,0.509,0.828}{48.6} & \cellcolor[rgb]{0.461,0.509,0.828}{46.1} & \cellcolor[rgb]{0.461,0.509,0.828}{35.0} & \cellcolor[rgb]{0.461,0.509,0.828}{47.0} & \cellcolor[rgb]{0.474,0.528,0.844}{62.4} & \cellcolor[rgb]{0.461,0.509,0.828}{48.1} \\
 &  &  & 64 & \cellcolor[rgb]{0.480,0.538,0.853}{40.2} & \cellcolor[rgb]{0.477,0.533,0.849}{43.8} & \cellcolor[rgb]{0.493,0.557,0.869}{45.5} & \cellcolor[rgb]{0.461,0.509,0.828}{48.6} & \cellcolor[rgb]{0.470,0.523,0.840}{46.0} & \cellcolor[rgb]{0.464,0.514,0.832}{34.7} & \cellcolor[rgb]{0.464,0.514,0.832}{46.9} & \cellcolor[rgb]{0.477,0.533,0.849}{62.3} & \cellcolor[rgb]{0.461,0.509,0.828}{48.0} \\
 &  &  & 128 & \cellcolor[rgb]{0.480,0.538,0.853}{40.2} & \cellcolor[rgb]{0.503,0.571,0.880}{43.6} & \cellcolor[rgb]{0.537,0.617,0.916}{45.1} & \cellcolor[rgb]{0.510,0.580,0.888}{48.1} & \cellcolor[rgb]{0.513,0.585,0.892}{45.6} & \cellcolor[rgb]{0.561,0.648,0.937}{32.1} & \cellcolor[rgb]{0.470,0.523,0.840}{46.8} & \cellcolor[rgb]{0.635,0.732,0.981}{59.0} & \cellcolor[rgb]{0.565,0.653,0.940}{46.0} \\
 &  &  & 256 & \cellcolor[rgb]{0.480,0.538,0.853}{40.2} & \cellcolor[rgb]{0.572,0.661,0.945}{43.1} & \cellcolor[rgb]{0.605,0.699,0.966}{44.5} & \cellcolor[rgb]{0.510,0.580,0.888}{48.1} & \cellcolor[rgb]{0.558,0.644,0.934}{45.2} & \cellcolor[rgb]{0.707,0.799,0.999}{28.7} & \cellcolor[rgb]{0.510,0.580,0.888}{46.4} & \cellcolor[rgb]{0.810,0.872,0.981}{55.5} & \cellcolor[rgb]{0.711,0.803,0.999}{43.5} \\
\midrule
LMK & Sentence & LMK & Sentence & \cellcolor[rgb]{0.480,0.538,0.853}{40.2} & \cellcolor[rgb]{0.503,0.571,0.880}{43.6} & \cellcolor[rgb]{0.503,0.571,0.880}{45.4} & \cellcolor[rgb]{0.500,0.566,0.876}{48.2} & \cellcolor[rgb]{0.503,0.571,0.880}{45.7} & \cellcolor[rgb]{0.734,0.821,0.999}{28.0} & \cellcolor[rgb]{0.896,0.904,0.917}{42.7} & \cellcolor[rgb]{0.464,0.514,0.832}{62.6} & \cellcolor[rgb]{0.658,0.755,0.990}{44.4} \\

\bottomrule
\end{tabular}
}
\caption{Comparison of pooling strategies across in-domain and out-of-domain retrieval benchmarks for \emph{ModernBERT-base} fine-tuned on MSMarco passage data. We report the pooling strategy and granularity (gran.) used during both fine-tuning and evaluation.}
\label{tab:modernbert_en_lmk_pooling_results}
\end{table*}

\section{Results \& Discussion}
\label{results}

% \todo[inline]{Claim 1) CLS pooling learns positional bias which is partly a data artefact but also arises from the modeling of how the embeddings are generated. Claim 2) This bias cannot be mitigated even after training with longer context, extrapolation ability is still worse. Claim 3) CLS vs Mean pooling face an inflection point where mean pooling starts to better extrapolate and CLS pooling extrapolation becomes worse (empirical) Claim 4) LMK pooling reduces this bias, empirical and supporting eqs from self attention which leads to better extrapolation. Claim 5) This pooling mechanism is resilient to different downstream tasks (empirical). TODOs: Embedding measures, Paired statistical significance tests for appendix, Train vs Eval different seq len comparisons on same data and HPs, LMK proximity and affinity results (what lmk is learning), anything else to support the claims?}

We aim to answer the following research questions: 1) How does \lmk pooling compare with other pooling strategies across short- and long-context tasks, and out-of-domain settings? 2) Is the positional bias learned by \texttt{CLS} pooling a modeling limitation or a data artefact? 3) Can training with longer contexts mitigate this bias? 4) Is \lmk pooling subject to positional bias, and how does it generalize across different downstream tasks?

\subsection{LMK Pooling Performance}
\label{subsec:lmk_results}

% We begin by examining English retrieval performance by training on the widely used MSMarco Passage dataset, as shown in \cref{tab:modernbert_en_lmk_pooling_results} for \emph{ModernBERT-base} and in \cref{tab:gte_en_lmk_pooling_results} (Appendix) for \emph{gte-en-mlm-base}. We observe that in-domain performance on the MSMarco Dev set with \lmk pooling is marginally better than all existing pooling baselines across different splitting strategies. For short-context tasks, LMK and CLS pooling perform comparably, while Mean and Latent attention pooling degrade performance by at least $\sim1\!$ points. This suggests that, for short-context tasks where CLS is commonly preferred, LMK pooling is also a strong alternative. In contrast, for long-context tasks, LMK pooling yields substantially larger gains compared to all other baselines. While Mean pooling improves over CLS on MLDR and LongEmbed, LMK pooling outperforms all methods by a significant margin. Finally, we find that using \emph{variable splitting} during LMK training generalizes well and provides flexibility in choosing the pooling granularity at inference time.We further report results on the full MTEB-v2 embedding benchmark in \cref{tab:mtebv2_gte_modernbert_msmarco_psg} (Appendix), where LMK pooling shows comparable performance to existing methods, consistent with the fact that MTEB-v2 focuses on short-context evaluation and does not assess realistic long-context behavior.

We examine English retrieval performance by traning on the MSMarco Passage dataset, showing results for \emph{ModernBERT-base} and \emph{gte-en-mlm-base} in \cref{tab:modernbert_en_lmk_pooling_results} and \cref{tab:gte_en_lmk_pooling_results} (Appendix) respectively. We observe that \lmk pooling peforms marginally better than all pooling baselines on the in-domain MSMarco Dev set. For short-context tasks, \lmk pooling maintains comparable performance to \texttt{CLS} pooling, both outperforming Mean and Latent attention pooling, indicating that \lmk is a strong alternative to \texttt{CLS} for such tasks. 
For long-context tasks, \lmk pooling significantly outperforms all other baselines. Moreover, we find that using \emph{Variable} splitting during \lmk training generalizes well while providing the flexibility of choosing the pooling granularity at inference time. 

% We additionally present results in the multilingual setting to assess whether LMK pooling generalizes across languages. We report performance for \emph{mmBERT-base} after standard contrastive finetuning on widely used multilingual retrieval benchmarks, as well as on a multilingual long-document classification benchmark, as shown in \cref{tab:mmbert_main_table_results}. We observe that LMK pooling yields substantial improvements on long-context evaluations both in and out-of-domain even in the multilingual setting. Full results are provided in \cref{app:additional_results}.

To assess whether \lmk pooling generalizes across languages, we report the performance of \emph{mmBERT-base} on widely uses multilingual retrieval and long-context classification benchmarks after contrastive finetuning. \cref{tab:mmbert_main_table_results} shows that \lmk pooling maintains substantial improvements on long-context evaluations across in and out-of-domain multilingual tasks. Full results are provided in \cref{app:additional_results}.

\begin{table}[ht]
\centering
\small
\setlength{\tabcolsep}{6pt}
\renewcommand{\arraystretch}{1.15}

\resizebox{\linewidth}{!}{%
\begin{tabular}{l c c c c c c}
\toprule

\multicolumn{1}{c}{\textsc{Pool}} &
\multicolumn{1}{c}{\textsc{FT.}} &
\multicolumn{1}{c}{\textsc{Eval.}} &
\multicolumn{2}{c}{\textsc{In-Domain}} &
\multicolumn{2}{c}{\textsc{Out-of-Domain}} \\

\cmidrule(lr){1-1}
\cmidrule(lr){2-2}
\cmidrule(lr){3-3}
\cmidrule(lr){4-5}
\cmidrule(lr){6-7}

&
\textbf{Gran.} &
\textbf{Gran.} &
\textbf{MIRACL} &
\textbf{MLDR} &
\textbf{M-Eurlex} &
\textbf{LongEmbed} \\

& & &
\textbf{(18)} &
\textbf{(13)} &
\textbf{(23)} &
\textbf{(6)} \\

 & & &
\texttt{NDCG@10} &
\texttt{NDCG@10} &
\texttt{F1} &
\texttt{P@1\;|\;NDCG@10} \\

\midrule

CLS     & --       & --       & 57.6 & 24.9 & 27.5 & 44.1 \\
Mean    & --       & --       & 56.1 & 17.0 & 27.5 & 48.4 \\
Latent  & --       & --       & 56.6 & 19.6 & 28.0 & 46.9 \\

\midrule

LMK     & 128      & 128      & \textbf{58.0} & 37.2 & 31.5 & 62.1 \\

\midrule

\multirow{4}{*}{LMK}
        & \multirow{4}{*}{Variable}
        & 32       & 57.8 & \textbf{38.7} & \textbf{33.7} & \textbf{70.7} \\
        &          & 64       & 57.9 & 38.6 & 33.4 & 70.7 \\
        &          & 128      & 57.9 & 38.3 & 33.2 & 70.2 \\
        &          & 256      & 57.7 & 38.0 & 33.0 & 69.8 \\

\bottomrule
\end{tabular}
}
\caption{Comparison of pooling strategies across in-domain and out-of-domain retrieval benchmarks for \emph{mmBERT-base} when fine-tuned with multilingual BGE-m3 training data.}
\label{tab:mmbert_main_table_results}
\end{table}

\begin{table*}[ht]
\centering
\small
\setlength{\tabcolsep}{5pt}
\renewcommand{\arraystretch}{1.15}

\resizebox{\textwidth}{!}{%
\begin{tabular}{c l c l c
                c
                c c c c
                c c c c}
\toprule

 & & & & &
\multicolumn{1}{c}{\textsc{In-Domain}} &
\multicolumn{4}{c}{\textsc{Out-of-Domain (Short Context)}} &
\multicolumn{4}{c}{\textsc{Out-of-Domain (Long Context)}} \\

\cmidrule(lr){6-6}
\cmidrule(lr){7-10}
\cmidrule(lr){11-14}

\textbf{Model} & & & & &
\textbf{MSMarco} &
\textbf{BEIR} &
\textbf{MTEBv2} &
\textbf{MIRACL} &
\textbf{Avg.} &
\textbf{MLDR} &
\textbf{COIR} &
\textbf{LongEmbed} &
\textbf{Avg.} \\

& \multicolumn{2}{c}{\textsc{Finetuning}} &
\multicolumn{2}{c}{\textsc{Evaluation}} &
\textbf{Dev} &
\textbf{(15)} &
\textbf{(10)} &
\textbf{Dev} &
\textbf{Short} &
\textbf{Test} &
\textbf{(8)} &
\textbf{(6)} &
\textbf{Long} \\

\cmidrule(lr){2-3}
\cmidrule(lr){4-5}

& \textbf{Pool} &
\textbf{Gran.} &
\textbf{Pool} &
\textbf{Gran.} &
\texttt{NDCG@10} &
\texttt{NDCG@10} &
\texttt{NDCG@10} &
\texttt{NDCG@10} &
 &
\texttt{NDCG@10} &
\texttt{NDCG@10} &
\texttt{P@1\;|\;NDCG@10} &
 \\

\midrule

\multirow{3}{*}{\emph{gte-en-mlm-base}}
& CLS & - & CLS & - & 28.7 & \textbf{37.6} & \textbf{39.0} & 32.3 & \textbf{36.3} & 29.1 & 30.1 & 49.1 & 36.1 \\
& Mean & - & Mean & - & 28.3 & 36.3 & 37.7 & 29.0 & 34.3 & 30.0 & \textbf{32.5} & \textbf{61.6} & \textbf{41.4} \\
& \multirow{1}{*}{LMK}
& \multirow{1}{*}{Variable}
& \multirow{1}{*}{LMK} & 32
& 28.9 & 37.1 & 38.1 & 32.1 & 35.8 & \textbf{31.4} & 30.9 & 59.8 & 40.7 \\

\midrule

\multirow{3}{*}{\emph{ModernBERT-base}}
& CLS & - & CLS & - & 26.0 & 38.0 & 39.8 & 32.0 & 36.6 & 29.7 & 44.2 & 58.3 & 44.1 \\
& Mean & - & Mean & - & 26.1 & 37.1 & 38.4 & 31.4 & 35.6 & 31.3 & 42.3 & \textbf{62.4} & \textbf{45.3} \\
& \multirow{1}{*}{LMK}
& \multirow{1}{*}{Variable}
& \multirow{1}{*}{LMK} & 32
& 26.0 & \textbf{38.7} & \textbf{40.4} & \textbf{33.1} & \textbf{37.4} & \textbf{32.2} & \textbf{45.6} & 58.1 & \textbf{45.3} \\

\bottomrule
\end{tabular}
}
\caption{Comparison of pooling strategies across in-domain and out-of-domain retrieval benchmarks for \emph{gte-en-mlm-base} and \emph{ModernBERT-base} when fine-tuned with MSMarco document data.}
\label{tab:merged_en_msmarco_doc_lmk_pooling_results}
\end{table*}

\subsection{Long-Context Performance: Data Artefacts or Modeling Limitations?}
\label{subsec:long_ctx_limitation_discussion}

% Having established that LMK pooling is effective for long-context generalization, we next examine whether this benefit persists when the training data itself contains very long documents, thereby reducing long-context bias. To this end, we use the MSMarco Document Ranking dataset and the multilingual BGE-m3 training dataset, increasing the training context length to 8k. \cref{tab:merged_en_msmarco_doc_lmk_pooling_results} reports results for two models trained under this long-context setting.

Having demonstrated the effectiveness of \lmk pooling in long-context generalization, we evaluate whether this persists when any context-length bias of the training data is reduced by adding very long documents to the training mix. Specifically, we use the MSMarco Document Ranking dataset, increase the training context length to 8k, and report results for two models trained in this long-context setting in \cref{tab:merged_en_msmarco_doc_lmk_pooling_results}.

% For \emph{gte-en-mlm-base}, \texttt{CLS} pooling achieves the strongest performance on short-context tasks, with LMK pooling performing comparably, while Mean pooling degrades performance by an additional $\sim\!1.5$ points. In contrast, on long-context tasks, Mean pooling outperforms both CLS and LMK, with CLS lagging substantially behind. For \emph{ModernBERT-base}, LMK pooling emerges as the most consistent method, achieving strong performance on both short- and long-context tasks and remaining competitive with both CLS and Mean pooling. Although CLS performs similarly to LMK on LongEmbed, closer inspection reveals that this improvement is driven primarily by the Passkey Retrieval dataset; on other long-context benchmarks, CLS continues to underperform, resulting in comparable macro-averaged scores.

For \emph{gte-en-mlm-base}, \texttt{CLS} pooling shows the strongest performance on short-context tasks, with \lmk pooling yielding comparable results. In contrast, on long-context tasks, Mean pooling outperforms both \texttt{CLS} and \lmk, with \texttt{CLS} substantially underperforming. For \emph{ModernBERT-base}, \lmk pooling is the most consistent method, performing well across both short- and long-context tasks. \texttt{CLS} achieves comparable scores on LongEmbed but primarily due to the Passkey Retrieval dataset; on other long-context benchmarks, \texttt{CLS} substantially underperforms (\cref{tab:modernbert_msmarco_doc_long_embed_results}).

\begin{figure}[ht]
  \centering
  \includegraphics[width=0.9\linewidth]{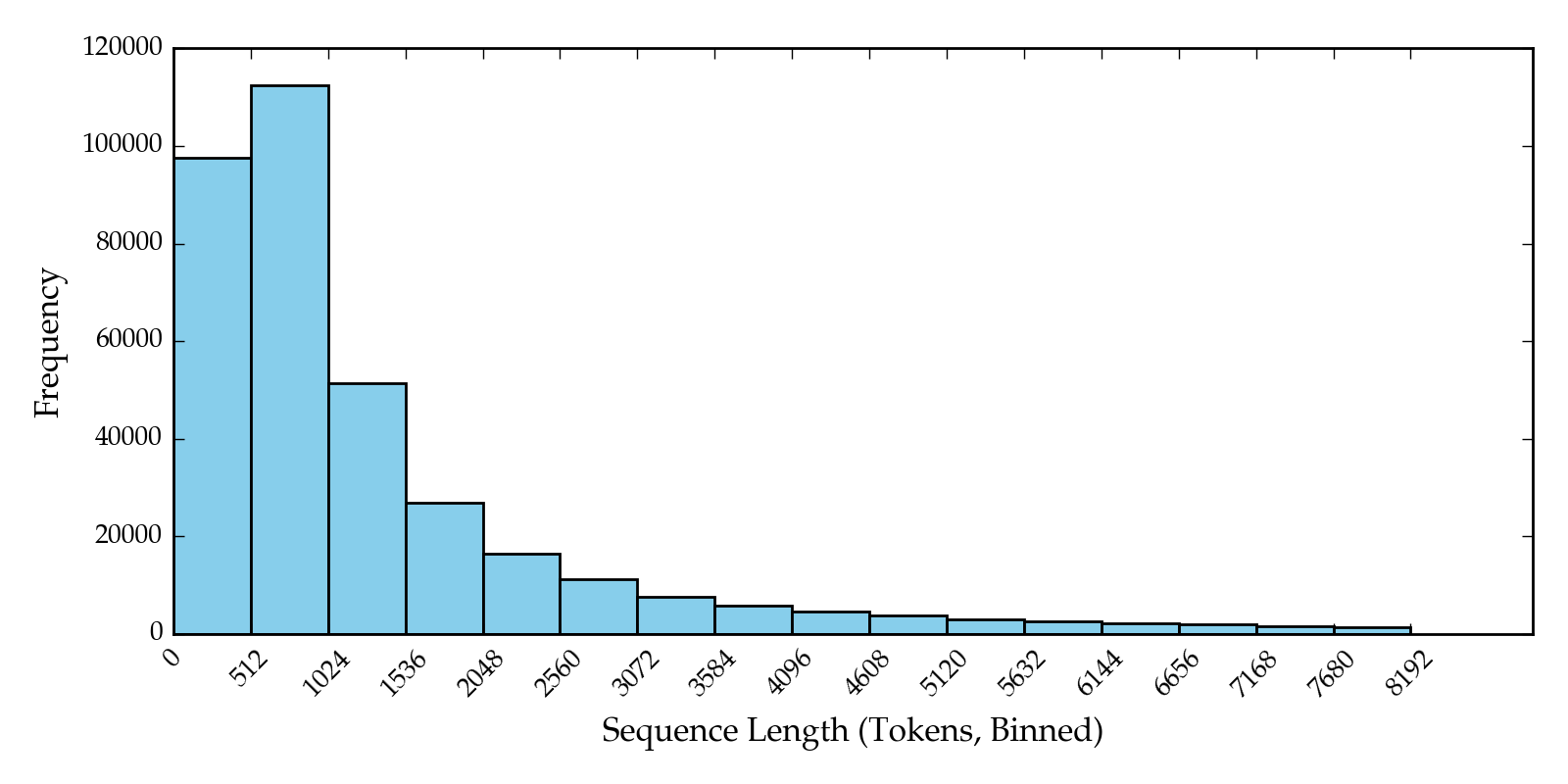}
\caption{Binned distribution of passage token lengths in the MS MARCO Document Ranking training set, computed using the \emph{ModernBERT-base} tokenizer.}
  \label{fig:msmarco_doc_length_plot}
\end{figure}

Although the MSMarco document ranking dataset is a popular long-context training source, its passages are not sufficiently long to enable robust long-context learning (\cref{fig:msmarco_doc_length_plot}). We instead finetune on the same multilingual training data employed in our earlier experiments, which includes the MLDR training set with significantly longer documents. All models are finetuned with a maximum sequence length of 8192 and a batch size of 64 for 5k steps. We vary the training sequence length for both \texttt{CLS} and \lmk pooling using \emph{mmBERT-base}, and report results in \cref{tab:mmbert_long_ctx_cls_lmk_comparison}. As training sequence length increases, \texttt{CLS} and \lmk achieve comparable performance on MLDR suggesting that \texttt{CLS} can partially mitigate long-context bias with longer training sequences. However, this does not generalize to out-of-domain datasets such as Multi-EURLEX and LongEmbed, where a substantial performance gap persists.

We conduct a similar experiment on \emph{gte-multilingual-base}, (\cref{tab:gte_long_ctx_ft_eval_pooling}), and observe a consistent trend: while CLS, Mean, and Latent Attention pooling become competitive with \lmk on MLDR, they substantially under-perform on out-of-domain long-context tasks. These results suggest indicate that long-context training can reduce but not eliminate the positional bias of these pooling methods.

\begin{table}[ht]
\centering
\small
\setlength{\tabcolsep}{6pt}
\renewcommand{\arraystretch}{1.15}

\resizebox{\linewidth}{!}{%
\begin{tabular}{c
                cc
                cc
                cc}
\toprule

 & \multicolumn{2}{c}{\textsc{In-Domain}} &
   \multicolumn{4}{c}{\textsc{Out-of-Domain}} \\

\cmidrule(lr){2-3}
\cmidrule(lr){4-7}

\textbf{FT Train} &
\multicolumn{2}{c}{\textbf{MLDR (13)}} &
\multicolumn{2}{c}{\textbf{Multi-EURLEX (23)}} &
\multicolumn{2}{c}{\textbf{LongEmbed (6)}} \\

\cmidrule(lr){2-3}
\cmidrule(lr){4-5}
\cmidrule(lr){6-7}

\textbf{MSL} &
\textbf{CLS} & \textbf{LMK} &
\textbf{CLS} & \textbf{LMK} &
\textbf{CLS} & \textbf{LMK} \\

 & \multicolumn{2}{c}{\texttt{NDCG@10}} & \multicolumn{2}{c}{\texttt{F1 Score}} & \multicolumn{2}{c}{\texttt{P@1\;|\;NDCG@10}} \\

\midrule

512  & 31.7 & \textbf{36.6} & 30.5 & \textbf{32.9} & 46.1 & \textbf{70.6} \\
1024 & 32.5 & \textbf{40.9} & 30.1 & \textbf{33.1} & 45.4 & \textbf{71.1} \\
2048 & 39.8 & \textbf{43.6} & 30.0 & \textbf{32.0} & 52.2 & \textbf{72.6 }\\
4096 & \textbf{46.3} & 45.9 & 29.4 & \textbf{31.7} & 52.5 & \textbf{73.3} \\
8192 & 49.1 & \textbf{49.6} & 30.2 & \textbf{31.5} & 56.2 & \textbf{71.3} \\

\bottomrule
\end{tabular}
}

\caption{Effect of fine-tuning maximum sequence length (MSL) on long-context embedding tasks for \emph{mmBERT-base}, comparing \texttt{CLS} and \lmk pooling.}
\label{tab:mmbert_long_ctx_cls_lmk_comparison}
\end{table}

\begin{table}[ht]
\centering
\small
\setlength{\tabcolsep}{6pt}
\renewcommand{\arraystretch}{1.15}

\resizebox{\linewidth}{!}{%
\begin{tabular}{l c c c c c c}
\toprule

\multicolumn{1}{c}{\textsc{Pool}} &
\multicolumn{1}{c}{\textsc{FT.}} &
\multicolumn{1}{c}{\textsc{Eval.}} &
\multicolumn{2}{c}{\textsc{In-Domain}} &
\multicolumn{2}{c}{\textsc{Out-of-Domain}} \\

\cmidrule(lr){1-1}
\cmidrule(lr){2-2}
\cmidrule(lr){3-3}
\cmidrule(lr){4-5}
\cmidrule(lr){6-7}

&
\textbf{Gran.} &
\textbf{Gran.} &
\textbf{MIRACL} &
\textbf{MLDR} &
\textbf{M-Eurlex} &
\textbf{LongEmbed} \\

& & &
\textbf{(18)} &
\textbf{(13)} &
\textbf{(23)} &
\textbf{(6)} \\

& & &
\texttt{NDCG@10} &
\texttt{NDCG@10} &
\texttt{F1} &
\texttt{P@1\;|\;NDCG@10} \\

\midrule

CLS     & --       & --       & \textbf{66.7} & \textbf{58.7} & 13.9 & 66.1 \\
Mean    & --       & --       & 66.3 & 58.3 & 14.5 & 65.8 \\
Latent  & --       & --       & 66.2 & 58.4 & 14.6 & 66.9 \\

\midrule

LMK     & 32       & 32       & 66.1 & 58.1 & 16.4 & 72.9 \\
LMK     & Variable & 32       & 66.2 & 58.1 & \textbf{18.0} & \textbf{73.1} \\

\bottomrule
\end{tabular}
}
\caption{Long-context finetuning ablations on \emph{gte-multilingual-base} with a maximum sequence length of 8192.}
\label{tab:gte_long_ctx_ft_eval_pooling}
\end{table}

\begin{table*}[ht]
\centering
\small
\setlength{\tabcolsep}{4pt}
\resizebox{\textwidth}{!}{%
\begin{tabular}{c l c l c
                c
                c c c c
                c c c c}
\toprule

  & & & & &
\multicolumn{1}{c}{\textsc{In-Domain}} &
\multicolumn{4}{c}{\textsc{Out-of-Domain (Short Context)}} &
\multicolumn{4}{c}{\textsc{Out-of-Domain (Long Context)}} \\

\cmidrule(lr){6-6}
\cmidrule(lr){7-10}
\cmidrule(lr){11-14}

\textbf{Model} & & & & &
\multicolumn{1}{c}{\textbf{MSMarco}} &
\textbf{BEIR} &
\textbf{MTEBv2} &
\textbf{MIRACL-HN} &
\textbf{Avg.} &
\textbf{MLDR} &
\textbf{COIR} &
\textbf{LongEmbed} &
\textbf{Avg.} \\

 & \multicolumn{2}{c}{\textsc{Finetuning}} &
\multicolumn{2}{c}{\textsc{Evaluation}} &
\multicolumn{1}{c}{\textbf{Dev}} &
\textbf{(15)} &
\textbf{(10)} &
\textbf{EN} &
\textbf{Short} &
\textbf{EN} &
\textbf{(8)} &
\textbf{(6)} &
\textbf{Long} \\

\cmidrule(lr){2-3}
\cmidrule(lr){4-5}

\textbf{} &
\textbf{Pool} &
\textbf{Gran.} &
\textbf{Pool} &
\textbf{Gran.} &
\texttt{NDCG@10} &
\texttt{NDCG@10} &
\texttt{NDCG@10} &
\texttt{NDCG@10} &
 &
\texttt{NDCG@10} &
\texttt{NDCG@10} &
\texttt{P@1\;|\;NDCG@10} &
 \\

\midrule

\multirow{4}{*}{\emph{gte-en-mlm-base}}
& CLS & -- & CLS & -- 
& 42.3 & 44.0 & 44.7 & 47.9 & 45.5 & 27.3 & 36.6 & 52.0 & 38.6 \\

& \multicolumn{4}{l}{\quad\emph{+ CLS RetroMAE pretraining}} 
& 42.5 & \textbf{44.7} & \textbf{45.5} & 47.4 & 45.9 & 31.2 & 42.1 & 60.8 & 44.7 \\

\cmidrule(lr){2-14}

& LMK & Variable & LMK & 256 
& \textbf{43.0} & 44.2 & 44.8 & 48.1 & 45.7 & 31.7 & 37.6 & 65.2 & 44.8 \\

& \multicolumn{4}{l}{\quad\emph{+ LMK RetroMAE pretraining}} 
& 42.6 & 44.5 & 45.4 & \textbf{48.4} & \textbf{46.1} & \textbf{36.6} & \textbf{43.3} & \textbf{67.7} & \textbf{49.2} \\

\midrule

\multirow{4}{*}{\emph{ModernBERT-base}}
& CLS & -- & CLS & -- 
& 39.8 & 42.8 & 44.2 & \textbf{48.7} & 45.2 & 24.9 & 43.5 & 43.3 & 37.2 \\

& \multicolumn{4}{l}{\quad\emph{+ CLS RetroMAE pretraining}} 
& 39.8 & 42.8 & 44.4 & 47.9 & 45.0 & 24.9 & 44.3 & 42.1 & 37.1 \\

\cmidrule(lr){2-14}

& LMK & Variable & LMK & 32 
& \textbf{40.3} & \textbf{44.0} & \textbf{45.8} & 48.6 & \textbf{46.1} & \textbf{35.0} & \textbf{47.0} & 62.4 & \textbf{48.1} \\

& \multicolumn{4}{l}{\quad\emph{+ LMK RetroMAE pretraining}} 
& 40.2 & 41.3 & 42.5 & 47.8 & 43.9 & 31.8 & 46.2 & \textbf{66.4} & \textbf{48.1} \\

\bottomrule
\end{tabular}
}
\caption{Effect of RetroMAE pretraining followed by fine-tuning for CLS and LMK pooling across \emph{gte-en-mlm-base} and \emph{ModernBERT-base} models.}
\label{tab:lmk_cls_retromae_pretraining}
\end{table*}

\subsection{Extension to Retrieval Pretraining}
\label{subsec:lmk_retromae}

% \todo[inline]{Changing Table 6 S.T it compares CLS before and after RetroMAE pretraining and same for LMK best config and also add Modernbert RetroMAE results in the same table.}

Retrieval oriented pretraining, as proposed in RetroMAE \cite{retromae-v1, retromae-v2}, is a popular self-supervised training method to learn effective text representations without the use of labeled data. It is often an intermediate pretraining step to strengthen vector embeddings before large-scale contrastive learning \cite{awasthy2025graniteembeddingr2models, bge-m3, sturua2024jinaembeddingsv3}. The method employs a two-tier architecture, where a standard encoder processes masked input sequences while a lightweight, bottleneck decoder attempts to reconstruct the original text from the encoder representations. While the \texttt{CLS} representation is typically used as an input to the enhanced decoder, we show that \lmk pooling can be used instead, providing significant performance gains in downstream long context retrieval tasks. 

Specifically, we follow the pretraining method proposed in \citet{retromae-v1} using about 225k samples from BookCorpus, Wikipedia and StackExchange, training for 2 epochs using a learning rate of $2\times10^{-5}$ and a maximum sequence length of 8192 tokens. We use \lmk pooling instead of the traditional \texttt{CLS}, experimenting with different granularities of inserting \lmk tokens into the sequence, before mean pooling across these tokens to reconstruct the masked sentence with the decoder. \cref{tab:lmk_cls_retromae_pretraining} compares pretraining with \lmk pooling and \texttt{CLS} pooling after finetuning on MSMarco passage data. We observe consistent performance gains from \lmk pretraining on \emph{gte-en-mlm-base}, whereas for \emph{ModernBERT-base}, both pooling strategies yield comparable results with no clear advantage.

\subsection{What does LMK capture?}
\label{subsec:lmk_capture}

We examine what \lmk pooling learns from token embeddings. \lmk pooling partitions an input sequence into fixed size chunks, generating one \lmk representation per chunk. Each \lmk forms a contextualized representation influenced by the full document. To understand what information \lmk embeddings preserve locally, we encode each chunk independently using \lmk and compare these non-contextualized embeddings with the contextualized \lmk embeddings.

\begin{figure}[ht]
  \centering
  \includegraphics[width=0.7\linewidth]{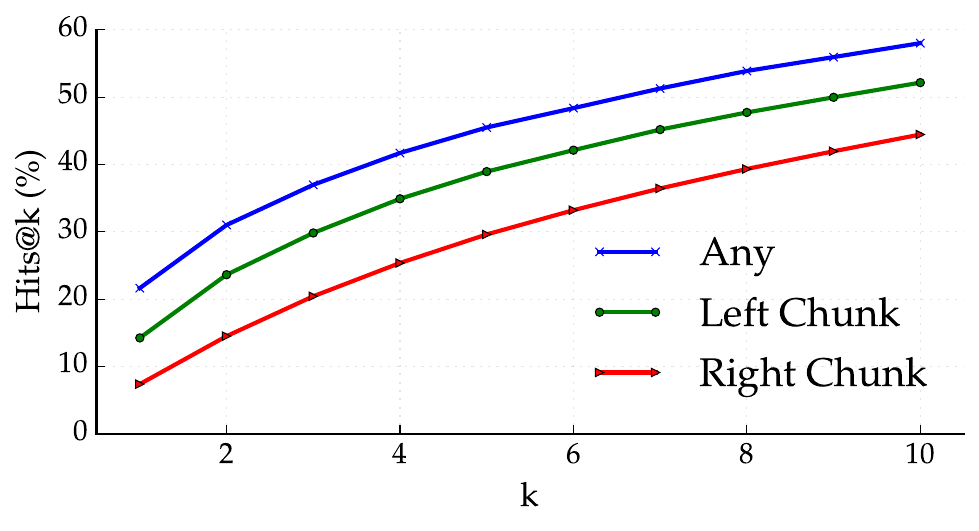}
  \caption{Retrieval Hits@k between LMK token embeddings and left, right, or any (left or right) chunk embeddings for sequences of length 8,192, with LMK granularity 32 ($\geq$248 chunks).}
  \label{fig:lmk_directional_hits}
\end{figure}

We sample long documents from the MLDR (En) dataset such that each document contains at least 256 chunks, and compute directional Hit@$k$ scores by retrieving the chunk embedding closest to each \lmk embedding, as shown in \cref{fig:lmk_directional_hits}. Since each \lmk token is positioned between a left and right chunk, we can measure directional retrieval bias.

We note two key findings. First, each left or right chunk embedding retrieves its corresponding \lmk embedding with $\ge 58\%$ Hit@10 accuracy, indicating that these representations preserve substantial local semantic information despite being contextualized by the entire document. Second, \lmk embeddings show higher affinity for left neighboring chunks than for right neighbors, indicating a directional bias- they emphasize earlier context more heavily than later context.

\section{Conclusion}
In this work, we first highlight biases and shortcomings of existing pooling mechanisms across various embedding tasks. We then introduce Landmark (\lmk) pooling, a simple yet effective method that partitions sequences into chunks, inserts landmark tokens, and mean-pools their embeddings. Extensive experiments across multiple encoder architectures, languages, and training setups show that \lmk pooling maintains competitive performance on short-context retrieval while providing substantial improvements on long-context evaluations, and enables robust extrapolation beyond training sequence lengths with minimal computational overhead. Further analysis of \lmk embeddings shows that they preserve local semantic information while incorporating global context, achieving an effective balance between representational capacity and discriminative power. The method integrates seamlessly with existing training and evaluation pipelines, offering a practical and robust alternative to prior approaches, with broad applicability to long-context embedding tasks, multi-vector embeddings, and chunking-based RAG systems.
\label{conclusion}

\section*{Impact Statement}

This paper presents work whose goal is to advance the field of 
Machine Learning. There are many potential societal consequences 
of our work, none which we feel must be specifically highlighted here.

\nocite{langley00}

\bibliography{example_paper, custom}
\bibliographystyle{icml2025}

%%%%%%%%%%%%%%%%%%%%%%%%%%%%%%%%%%%%%%%%%%%%%%%%%%%%%%%%%%%%%%%%%%%%%%%%%%%%%%%
%%%%%%%%%%%%%%%%%%%%%%%%%%%%%%%%%%%%%%%%%%%%%%%%%%%%%%%%%%%%%%%%%%%%%%%%%%%%%%%
% APPENDIX
%%%%%%%%%%%%%%%%%%%%%%%%%%%%%%%%%%%%%%%%%%%%%%%%%%%%%%%%%%%%%%%%%%%%%%%%%%%%%%%
%%%%%%%%%%%%%%%%%%%%%%%%%%%%%%%%%%%%%%%%%%%%%%%%%%%%%%%%%%%%%%%%%%%%%%%%%%%%%%%
% \newpage
\clearpage

\onecolumn

\appendix

\section{Additional Results}
\label{app:additional_results}

\begin{figure*}[ht]
  \centering

  % Row 1
  \subfigure[QMSum]{
    \includegraphics[width=0.48\textwidth]{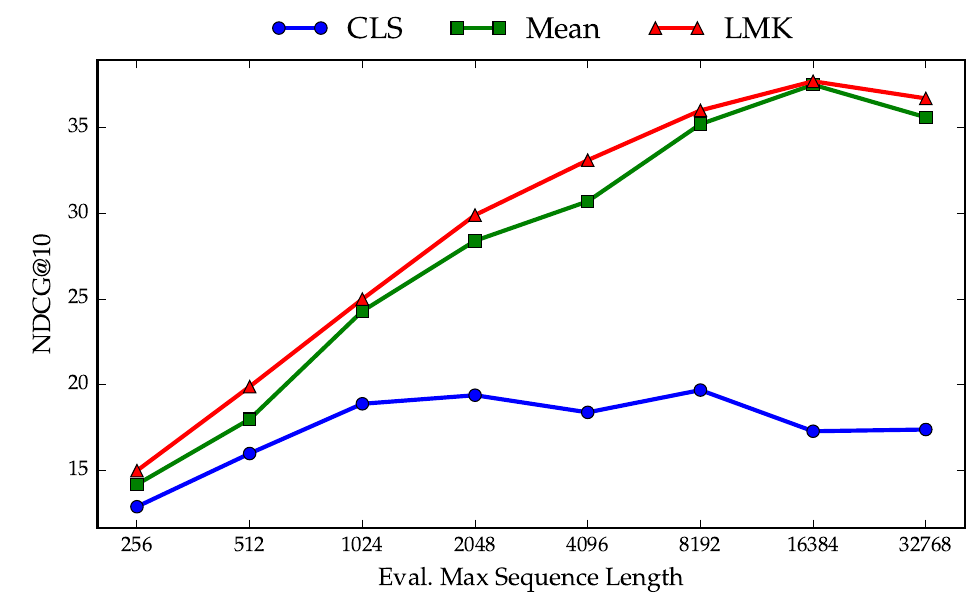}
    \label{fig:long_ctx_qmsum}
  }
  \hfill
  \subfigure[SummScreenFD]{
    \includegraphics[width=0.48\textwidth]{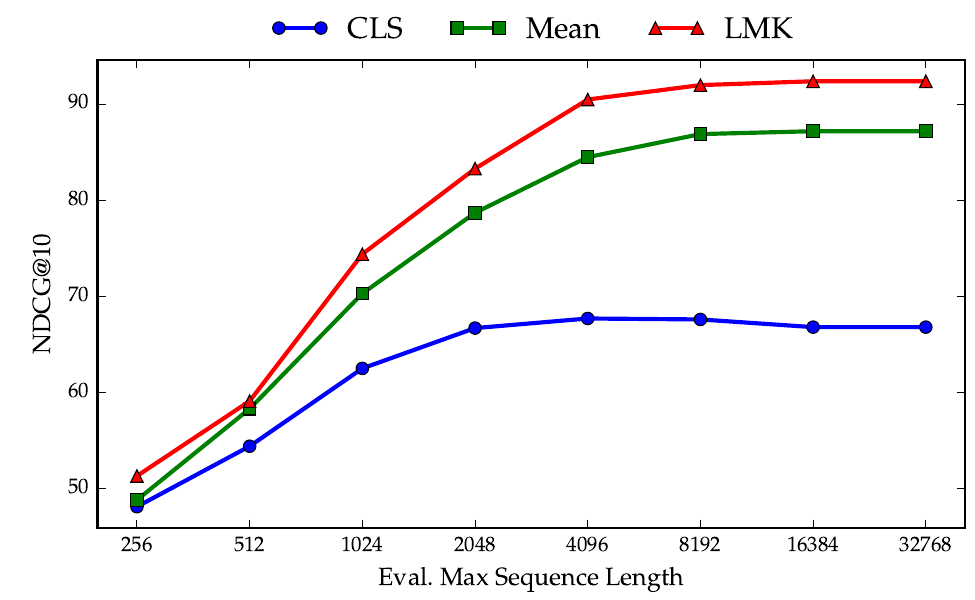}
    \label{fig:long_ctx_summscreenfd}
  }

  % \vspace{0.6em}

  % Row 2
  \subfigure[WikiMQA]{
    \includegraphics[width=0.48\textwidth]{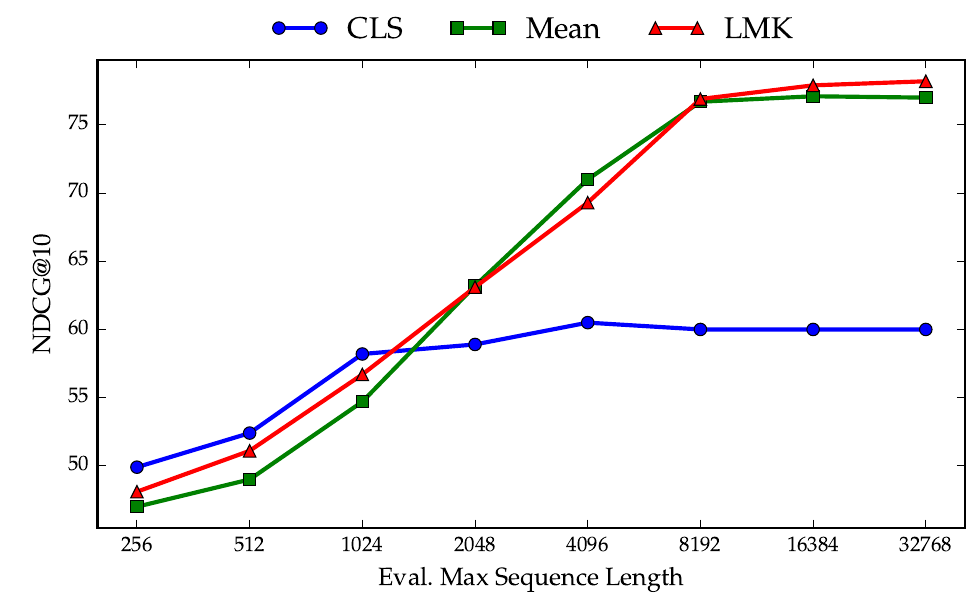}
    \label{fig:long_ctx_wikimqa}
  }
  \hfill
  \subfigure[NarrativeQA]{
    \includegraphics[width=0.48\textwidth]{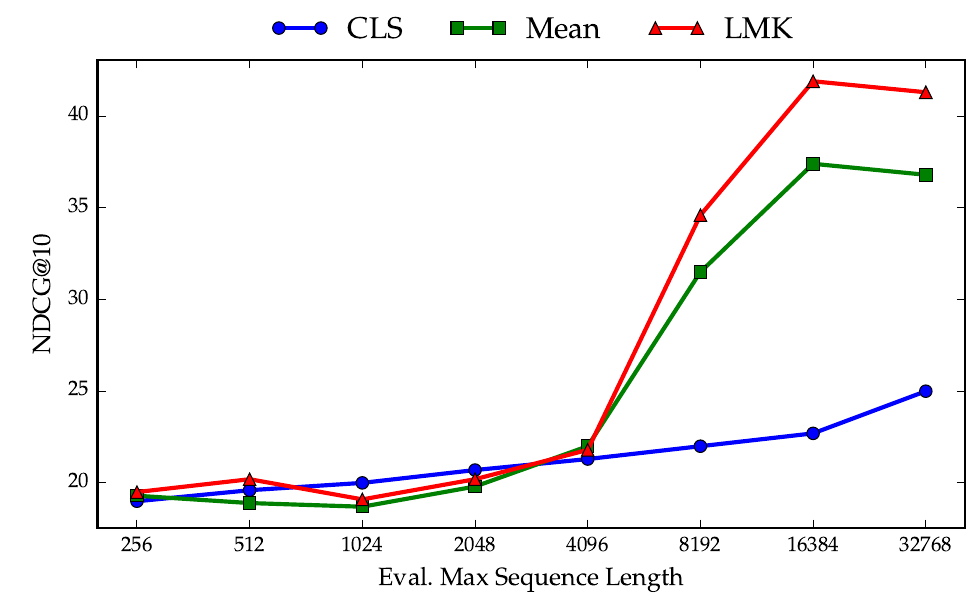}
    \label{fig:long_ctx_nqa}
  }

  % \vspace{0.6em}

  % Row 3
  \subfigure[Needle]{
    \includegraphics[width=0.48\textwidth]{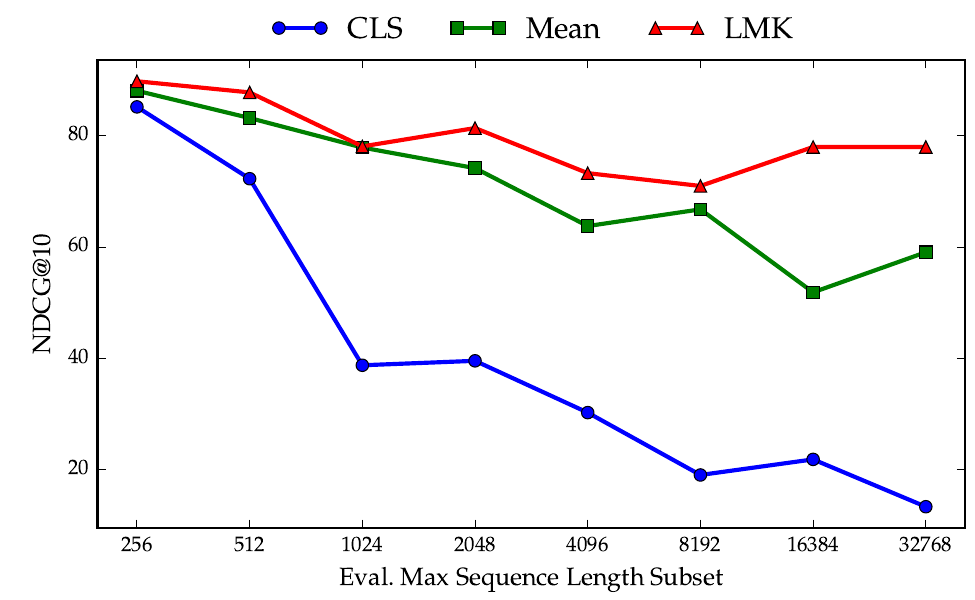}
    \label{fig:long_ctx_needle}
  }
  \hfill
  \subfigure[Passkey]{
    \includegraphics[width=0.48\textwidth]{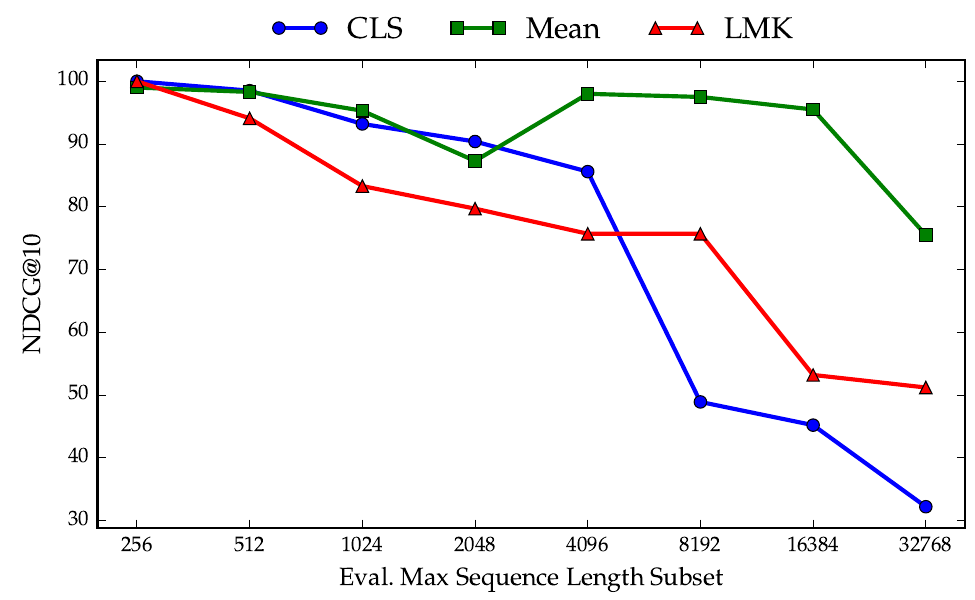}
    \label{fig:long_ctx_passkey}
  }

  \caption{Long-context retrieval performance across datasets for different pooling strategies using \emph{ModernBERT-base} fine-tuned on MSMarco Passages.}
  \label{fig:long_ctx_all}
\end{figure*}

In this section, we present additional results that complement the main findings. In \cref{tab:gte_en_lmk_pooling_results}, we report finetuning results on the MSMarco passage ranking dataset using \emph{gte-en-mlm-base}, analogous to the \emph{ModernBERT-base} results shown in \cref{tab:modernbert_en_lmk_pooling_results} in the main text. We observe similar trends, where \lmk pooling consistently exhibits strong long context extrapolation, and using variable granularity during finetuning provides added robustness when selecting granularity at inference time. We additionally illustrate the effects of long context extrapolation in \cref{fig:long_ctx_qmsum,fig:long_ctx_summscreenfd,fig:long_ctx_wikimqa,fig:long_ctx_nqa,fig:long_ctx_needle,fig:long_ctx_passkey}, showing how \lmk performance on LongEmbed datasets improves with increasing context length, while \texttt{CLS} pooling fails to effectively leverage the benefits of longer contexts. Extending to pretraining with RetroMAE, \cref{tab:gte_en_pretrain_results,tab:modernbert_en_pretrain_results} report the performance of both base models after RetroMAE pretraining with \texttt{CLS} versus \lmk pooling, indicating that such pretraining extensions can be readily applied to the proposed pooling mechanism.

We further report results for finetuning on the MSMarco document ranking dataset in \cref{tab:gte_en_msmarco_doc_lmk_pooling_results}. Here, \lmk pooling achieves the best overall in domain performance. For short context benchmarks, \texttt{CLS} pooling attains the highest average score, outperforming \lmk by $\sim\!0.5$ points, while mean pooling leads to a notable degradation. In long context evaluations, mean pooling performs best overall, with \lmk closely following, whereas \texttt{CLS} pooling lags substantially behind.

We additionally report the full multilingual results on MIRACL, MLDR, and Multi-EURLEX in \cref{tab:mmbert_miracl_results,tab:mmbert_mldr_results,tab:mmbert_multieurlex_results}, corresponding to \cref{tab:mmbert_main_table_results} in the main text.

\begin{table*}[ht]
\centering
\small
\setlength{\tabcolsep}{5pt}
\renewcommand{\arraystretch}{1.15}

\resizebox{\textwidth}{!}{%
\begin{tabular}{l c l c
                c
                c c c c
                c c c c}
\toprule

 & & & &
\multicolumn{1}{c}{\textsc{In-Domain}} &
\multicolumn{4}{c}{\textsc{Out-of-Domain (Short Context)}} &
\multicolumn{4}{c}{\textsc{Out-of-Domain (Long Context)}} \\

\cmidrule(lr){5-5}
\cmidrule(lr){6-9}
\cmidrule(lr){10-13}

& & & &
\textbf{MSMarco} &
\textbf{BEIR} &
\textbf{MTEBv2} &
\textbf{MIRACL} &
\textbf{Avg.} &
\textbf{MLDR} &
\textbf{COIR} &
\textbf{LongEmbed} &
\textbf{Avg.} \\

\multicolumn{2}{c}{\textsc{Finetuning}} &
\multicolumn{2}{c}{\textsc{Evaluation}} &
\textbf{Dev} &
\textbf{(15)} &
\textbf{(10)} &
\textbf{Dev} &
\textbf{Short} &
\textbf{Test} &
\textbf{(8)} &
\textbf{(6)} &
\textbf{Long} \\

\cmidrule(lr){1-2}
\cmidrule(lr){3-4}

\textbf{Pool} &
\textbf{Gran.} &
\textbf{Pool} &
\textbf{Gran.} &
\texttt{NDCG@10} &
\texttt{NDCG@10} &
\texttt{NDCG@10} &
\texttt{NDCG@10} &
 &
\texttt{NDCG@10} &
\texttt{NDCG@10} &
\texttt{P@1\;|\;NDCG@10} &
 \\

\midrule

CLS & -- & CLS & -- & \cellcolor[rgb]{0.978,0.776,0.694}{42.3} & \cellcolor[rgb]{0.569,0.657,0.942}{44.0} & \cellcolor[rgb]{0.587,0.679,0.955}{44.7} & \cellcolor[rgb]{0.561,0.648,0.937}{47.9} & \cellcolor[rgb]{0.551,0.635,0.928}{45.5} & \cellcolor[rgb]{0.903,0.554,0.503}{27.3} & \cellcolor[rgb]{0.794,0.311,0.405}{36.6} & \cellcolor[rgb]{0.794,0.311,0.405}{52.0} & \cellcolor[rgb]{0.794,0.311,0.405}{38.6} \\
CLS & -- & MultiCLS & 128 & \cellcolor[rgb]{0.680,0.776,0.995}{42.8} & \cellcolor[rgb]{0.864,0.482,0.461}{42.5} & \cellcolor[rgb]{0.794,0.311,0.405}{42.5} & \cellcolor[rgb]{0.794,0.311,0.405}{42.6} & \cellcolor[rgb]{0.794,0.311,0.405}{42.5} & \cellcolor[rgb]{0.824,0.879,0.974}{31.0} & \cellcolor[rgb]{0.970,0.719,0.634}{36.9} & \cellcolor[rgb]{0.870,0.898,0.943}{60.5} & \cellcolor[rgb]{0.854,0.893,0.956}{42.8} \\
Mean & -- & Mean & -- & \cellcolor[rgb]{0.906,0.560,0.507}{42.1} & \cellcolor[rgb]{0.810,0.361,0.417}{42.4} & \cellcolor[rgb]{0.976,0.818,0.746}{43.9} & \cellcolor[rgb]{0.976,0.818,0.746}{45.6} & \cellcolor[rgb]{0.977,0.764,0.681}{44.0} & \cellcolor[rgb]{0.774,0.851,0.994}{31.4} & \cellcolor[rgb]{0.794,0.311,0.405}{36.6} & \cellcolor[rgb]{0.847,0.890,0.960}{60.8} & \cellcolor[rgb]{0.838,0.886,0.967}{42.9} \\
Latent Attn & -- & Latent Attn & -- & \cellcolor[rgb]{0.827,0.410,0.429}{42.0} & \cellcolor[rgb]{0.794,0.311,0.405}{42.3} & \cellcolor[rgb]{0.977,0.760,0.677}{43.8} & \cellcolor[rgb]{0.945,0.644,0.568}{44.9} & \cellcolor[rgb]{0.936,0.624,0.553}{43.7} & \cellcolor[rgb]{0.907,0.905,0.904}{30.2} & \cellcolor[rgb]{0.934,0.619,0.549}{36.8} & \cellcolor[rgb]{0.774,0.851,0.994}{61.7} & \cellcolor[rgb]{0.838,0.886,0.967}{42.9} \\
Mean@k & 64 & Mean@k & 64 & \cellcolor[rgb]{0.794,0.311,0.405}{41.9} & \cellcolor[rgb]{0.941,0.884,0.852}{43.2} & \cellcolor[rgb]{0.879,0.901,0.935}{44.2} & \cellcolor[rgb]{0.803,0.868,0.984}{46.8} & \cellcolor[rgb]{0.854,0.893,0.956}{44.7} & \cellcolor[rgb]{0.969,0.843,0.780}{29.3} & \cellcolor[rgb]{0.957,0.866,0.819}{37.1} & \cellcolor[rgb]{0.911,0.571,0.514}{56.0} & \cellcolor[rgb]{0.930,0.608,0.541}{40.8} \\
\midrule

\multirow{4}{*}{LMK} & 32 & \multirow{4}{*}{LMK} & 32 & \cellcolor[rgb]{0.680,0.776,0.995}{42.8} & \cellcolor[rgb]{0.474,0.528,0.844}{44.2} & \cellcolor[rgb]{0.477,0.533,0.849}{44.9} & \cellcolor[rgb]{0.882,0.902,0.932}{46.4} & \cellcolor[rgb]{0.665,0.762,0.991}{45.2} & \cellcolor[rgb]{0.824,0.879,0.974}{31.0} & \cellcolor[rgb]{0.461,0.509,0.828}{38.0} & \cellcolor[rgb]{0.579,0.670,0.950}{63.9} & \cellcolor[rgb]{0.565,0.653,0.940}{44.3} \\
 & 64 &  & 64 & \cellcolor[rgb]{0.513,0.585,0.892}{43.0} & \cellcolor[rgb]{0.520,0.594,0.898}{44.1} & \cellcolor[rgb]{0.477,0.533,0.849}{44.9} & \cellcolor[rgb]{0.899,0.905,0.914}{46.3} & \cellcolor[rgb]{0.703,0.796,0.998}{45.1} & \cellcolor[rgb]{0.794,0.311,0.405}{24.7} & \cellcolor[rgb]{0.609,0.704,0.968}{37.6} & \cellcolor[rgb]{0.569,0.657,0.942}{64.0} & \cellcolor[rgb]{0.953,0.871,0.827}{42.1} \\
 & 128 &  & 128 & \cellcolor[rgb]{0.461,0.509,0.828}{43.2} & \cellcolor[rgb]{0.474,0.528,0.844}{44.2} & \cellcolor[rgb]{0.477,0.533,0.849}{44.9} & \cellcolor[rgb]{0.737,0.824,0.999}{47.1} & \cellcolor[rgb]{0.590,0.683,0.957}{45.4} & \cellcolor[rgb]{0.587,0.679,0.955}{32.9} & \cellcolor[rgb]{0.609,0.704,0.968}{37.6} & \cellcolor[rgb]{0.503,0.571,0.880}{64.8} & \cellcolor[rgb]{0.461,0.509,0.828}{45.1} \\
 & 256 &  & 256 & \cellcolor[rgb]{0.513,0.585,0.892}{43.0} & \cellcolor[rgb]{0.569,0.657,0.942}{44.0} & \cellcolor[rgb]{0.646,0.743,0.985}{44.6} & \cellcolor[rgb]{0.461,0.509,0.828}{49.0} & \cellcolor[rgb]{0.461,0.509,0.828}{45.9} & \cellcolor[rgb]{0.461,0.509,0.828}{34.6} & \cellcolor[rgb]{0.934,0.619,0.549}{36.8} & \cellcolor[rgb]{0.924,0.896,0.880}{59.7} & \cellcolor[rgb]{0.684,0.780,0.996}{43.7} \\
\midrule

\multirow{4}{*}{LMK}
& \multirow{4}{*}{Variable}
& \multirow{4}{*}{LMK} & 32 & \cellcolor[rgb]{0.513,0.585,0.892}{43.0} & \cellcolor[rgb]{0.569,0.657,0.942}{44.0} & \cellcolor[rgb]{0.587,0.679,0.955}{44.7} & \cellcolor[rgb]{0.824,0.879,0.974}{46.7} & \cellcolor[rgb]{0.703,0.796,0.998}{45.1} & \cellcolor[rgb]{0.945,0.644,0.568}{27.8} & \cellcolor[rgb]{0.533,0.612,0.912}{37.7} & \cellcolor[rgb]{0.486,0.547,0.861}{65.0} & \cellcolor[rgb]{0.722,0.812,1.000}{43.5} \\
 &  &  & 64 & \cellcolor[rgb]{0.594,0.687,0.960}{42.9} & \cellcolor[rgb]{0.569,0.657,0.942}{44.0} & \cellcolor[rgb]{0.587,0.679,0.955}{44.7} & \cellcolor[rgb]{0.882,0.902,0.932}{46.4} & \cellcolor[rgb]{0.745,0.830,0.999}{45.0} & \cellcolor[rgb]{0.948,0.878,0.840}{29.7} & \cellcolor[rgb]{0.688,0.783,0.997}{37.5} & \cellcolor[rgb]{0.480,0.538,0.853}{65.1} & \cellcolor[rgb]{0.605,0.699,0.966}{44.1} \\
 &  &  & 128 & \cellcolor[rgb]{0.513,0.585,0.892}{43.0} & \cellcolor[rgb]{0.569,0.657,0.942}{44.0} & \cellcolor[rgb]{0.587,0.679,0.955}{44.7} & \cellcolor[rgb]{0.760,0.840,0.996}{47.0} & \cellcolor[rgb]{0.665,0.762,0.991}{45.2} & \cellcolor[rgb]{0.844,0.889,0.962}{30.8} & \cellcolor[rgb]{0.688,0.783,0.997}{37.5} & \cellcolor[rgb]{0.461,0.509,0.828}{65.7} & \cellcolor[rgb]{0.493,0.557,0.869}{44.7} \\
 &  &  & 256 & \cellcolor[rgb]{0.513,0.585,0.892}{43.0} & \cellcolor[rgb]{0.474,0.528,0.844}{44.2} & \cellcolor[rgb]{0.530,0.608,0.909}{44.8} & \cellcolor[rgb]{0.520,0.594,0.898}{48.1} & \cellcolor[rgb]{0.483,0.543,0.857}{45.7} & \cellcolor[rgb]{0.737,0.824,0.999}{31.7} & \cellcolor[rgb]{0.609,0.704,0.968}{37.6} & \cellcolor[rgb]{0.474,0.528,0.844}{65.2} & \cellcolor[rgb]{0.477,0.533,0.849}{44.8} \\
\midrule
LMK & Sentence & LMK & Sentence & \cellcolor[rgb]{0.847,0.890,0.960}{42.6} & \cellcolor[rgb]{0.461,0.509,0.828}{44.3} & \cellcolor[rgb]{0.461,0.509,0.828}{45.0} & \cellcolor[rgb]{0.782,0.855,0.992}{46.9} & \cellcolor[rgb]{0.590,0.683,0.957}{45.4} & \cellcolor[rgb]{0.493,0.557,0.869}{33.7} & \cellcolor[rgb]{0.794,0.311,0.405}{36.6} & \cellcolor[rgb]{0.831,0.883,0.971}{61.0} & \cellcolor[rgb]{0.665,0.762,0.991}{43.8} \\

\bottomrule
\end{tabular}
}
\caption{Comparison of pooling strategies across in-domain and out-of-domain retrieval benchmarks for \emph{gte-en-mlm-base} when fine-tuned with MSMarco passage data.}
\label{tab:gte_en_lmk_pooling_results}
\end{table*}

\begin{table*}[ht]
\centering
\small
\setlength{\tabcolsep}{5pt}
\renewcommand{\arraystretch}{1.15}

\resizebox{\textwidth}{!}{%
\begin{tabular}{l c l c
                c
                c c c c
                c c c c}
\toprule

 & & & &
\multicolumn{1}{c}{\textsc{In-Domain}} &
\multicolumn{4}{c}{\textsc{Out-of-Domain (Short Context)}} &
\multicolumn{4}{c}{\textsc{Out-of-Domain (Long Context)}} \\

\cmidrule(lr){5-5}
\cmidrule(lr){6-9}
\cmidrule(lr){10-13}

& & & &
\textbf{MSMarco} &
\textbf{BEIR} &
\textbf{MTEBv2} &
\textbf{MIRACL} &
\textbf{Avg.} &
\textbf{MLDR} &
\textbf{COIR} &
\textbf{LongEmbed} &
\textbf{Avg.} \\

\multicolumn{2}{c}{\textsc{Finetuning}} &
\multicolumn{2}{c}{\textsc{Evaluation}} &
\textbf{Dev} &
\textbf{(15)} &
\textbf{(10)} &
\textbf{Dev} &
\textbf{Short} &
\textbf{Test} &
\textbf{(8)} &
\textbf{(6)} &
\textbf{Long} \\

\cmidrule(lr){1-2}
\cmidrule(lr){3-4}

\textbf{Pool} &
\textbf{Gran.} &
\textbf{Pool} &
\textbf{Gran.} &
\texttt{NDCG@10} &
\texttt{NDCG@10} &
\texttt{NDCG@10} &
\texttt{NDCG@10} &
 &
\texttt{NDCG@10} &
\texttt{NDCG@10} &
\texttt{P@1\;|\;NDCG@10} &
 \\

\midrule

CLS & - & CLS & - & \cellcolor[rgb]{0.726,0.815,1.000}{28.7} & \cellcolor[rgb]{0.461,0.509,0.828}{37.6} & \cellcolor[rgb]{0.461,0.509,0.828}{39.0} & \cellcolor[rgb]{0.461,0.509,0.828}{32.3} & \cellcolor[rgb]{0.461,0.509,0.828}{36.3} & \cellcolor[rgb]{0.794,0.311,0.405}{29.1} & \cellcolor[rgb]{0.794,0.311,0.405}{30.1} & \cellcolor[rgb]{0.794,0.311,0.405}{49.1} & \cellcolor[rgb]{0.794,0.311,0.405}{36.1} \\
Mean & - & Mean & - & \cellcolor[rgb]{0.794,0.311,0.405}{28.3} & \cellcolor[rgb]{0.794,0.311,0.405}{36.3} & \cellcolor[rgb]{0.903,0.554,0.503}{37.7} & \cellcolor[rgb]{0.794,0.311,0.405}{29.0} & \cellcolor[rgb]{0.794,0.311,0.405}{34.3} & \cellcolor[rgb]{0.975,0.821,0.750}{30.0} & \cellcolor[rgb]{0.461,0.509,0.828}{32.5} & \cellcolor[rgb]{0.544,0.626,0.922}{61.6} & \cellcolor[rgb]{0.486,0.547,0.861}{41.4} \\
\midrule

LMK & 128 & LMK & 128 & \cellcolor[rgb]{0.904,0.906,0.907}{28.6} & \cellcolor[rgb]{0.975,0.744,0.660}{36.7} & \cellcolor[rgb]{0.794,0.311,0.405}{37.5} & \cellcolor[rgb]{0.973,0.736,0.651}{30.1} & \cellcolor[rgb]{0.954,0.668,0.589}{34.8} & \cellcolor[rgb]{0.975,0.821,0.750}{30.0} & \cellcolor[rgb]{0.730,0.818,0.999}{31.6} & \cellcolor[rgb]{0.461,0.509,0.828}{63.1} & \cellcolor[rgb]{0.461,0.509,0.828}{41.6} \\
\midrule

\multirow{4}{*}{LMK}
& \multirow{4}{*}{Variable}
& \multirow{4}{*}{LMK} & 32 & \cellcolor[rgb]{0.461,0.509,0.828}{28.9} & \cellcolor[rgb]{0.771,0.848,0.994}{37.1} & \cellcolor[rgb]{0.932,0.891,0.868}{38.1} & \cellcolor[rgb]{0.500,0.566,0.876}{32.1} & \cellcolor[rgb]{0.642,0.739,0.984}{35.8} & \cellcolor[rgb]{0.461,0.509,0.828}{31.4} & \cellcolor[rgb]{0.971,0.840,0.776}{30.9} & \cellcolor[rgb]{0.711,0.803,0.999}{59.8} & \cellcolor[rgb]{0.635,0.732,0.981}{40.7} \\
 &  &  & 64 & \cellcolor[rgb]{0.977,0.764,0.681}{28.5} & \cellcolor[rgb]{0.930,0.893,0.872}{36.9} & \cellcolor[rgb]{0.977,0.764,0.681}{37.9} & \cellcolor[rgb]{0.851,0.891,0.958}{31.0} & \cellcolor[rgb]{0.904,0.906,0.907}{35.3} & \cellcolor[rgb]{0.547,0.630,0.925}{31.1} & \cellcolor[rgb]{0.966,0.705,0.622}{30.6} & \cellcolor[rgb]{0.654,0.751,0.988}{60.4} & \cellcolor[rgb]{0.635,0.732,0.981}{40.7} \\
 &  &  & 128 & \cellcolor[rgb]{0.977,0.764,0.681}{28.5} & \cellcolor[rgb]{0.771,0.848,0.994}{37.1} & \cellcolor[rgb]{0.971,0.840,0.776}{38.0} & \cellcolor[rgb]{0.821,0.878,0.976}{31.1} & \cellcolor[rgb]{0.860,0.895,0.951}{35.4} & \cellcolor[rgb]{0.904,0.906,0.907}{30.3} & \cellcolor[rgb]{0.919,0.587,0.526}{30.4} & \cellcolor[rgb]{0.692,0.786,0.997}{60.0} & \cellcolor[rgb]{0.745,0.830,0.999}{40.2} \\
 &  &  & 256 & \cellcolor[rgb]{0.977,0.764,0.681}{28.5} & \cellcolor[rgb]{0.930,0.893,0.872}{36.9} & \cellcolor[rgb]{0.977,0.764,0.681}{37.9} & \cellcolor[rgb]{0.972,0.834,0.768}{30.4} & \cellcolor[rgb]{0.969,0.843,0.780}{35.1} & \cellcolor[rgb]{0.954,0.668,0.589}{29.7} & \cellcolor[rgb]{0.794,0.311,0.405}{30.1} & \cellcolor[rgb]{0.703,0.796,0.998}{59.9} & \cellcolor[rgb]{0.810,0.872,0.981}{39.9} \\

\bottomrule
\end{tabular}
}
\caption{Comparison of pooling strategies across in-domain and out-of-domain retrieval benchmarks for \emph{gte-en-mlm-base} when fine-tuned with MSMarco document data.}
\label{tab:gte_en_msmarco_doc_lmk_pooling_results}
\end{table*}

\begin{table*}[ht]
\centering
\small
\setlength{\tabcolsep}{5pt}
\renewcommand{\arraystretch}{1.15}

\resizebox{0.9\textwidth}{!}{%
\begin{tabular}{c c c c
                c c c c
                c c
                c}
\toprule

\multicolumn{2}{c}{\textsc{Finetuning}} &
\multicolumn{2}{c}{\textsc{Evaluation}} &
\multicolumn{7}{c}{\textsc{Long-Embed Retrieval}}
\\

\cmidrule(lr){1-2}
\cmidrule(lr){3-4}
\cmidrule(lr){5-11}

\textbf{Pool} &
\textbf{Gran.} &
\textbf{Pool} &
\textbf{Gran.} &
\textbf{NarrativeQA} &
\textbf{QMSum} &
\textbf{WikiMQA} &
\textbf{SummScreenFD} &
\textbf{Needle} &
\textbf{Passkey} &
\textbf{Avg.} \\

& & & &
\texttt{NDCG@10} &
\texttt{NDCG@10} &
\texttt{NDCG@10} &
\texttt{NDCG@10} &
\texttt{P@1} &
\texttt{P@1} &
\\

\midrule

CLS  & --       & CLS  & -- & 30.9 & 32.3 & 75.0 & 82.4 & 46.5 & \textbf{82.8} & 58.3 \\
Mean & --       & Mean & -- & \textbf{47.1} & \textbf{35.7} & \textbf{81.1} & 86.8 & \textbf{48.3} & 75.8 & \textbf{62.4} \\
LMK  & Variable & LMK  & 32 & 44.4 & 34.7 & 79.8 & \textbf{89.5} & 41.8 & 58.5 & 58.1 \\

\bottomrule
\end{tabular}
}
\caption{Comparison of pooling strategies on LongEmbed retrieval benchmark for \emph{modernbert-base} fine-tuned on MSMarco document data.}
\label{tab:modernbert_msmarco_doc_long_embed_results}
\end{table*}

\begin{table*}[ht]
\centering
\small
\setlength{\tabcolsep}{3.5pt}
\renewcommand{\arraystretch}{1.15}

\resizebox{\textwidth}{!}{%
\begin{tabular}{l c l c
ccccccccccccccccccc}
\toprule

\textbf{FT. Pool} &
\textbf{FT. Gran.} &
\textbf{Eval. Pool} &
\textbf{Eval. Gran.} &
\textbf{ara} &
\textbf{ben} &
\textbf{deu} &
\textbf{eng} &
\textbf{fas} &
\textbf{fin} &
\textbf{fra} &
\textbf{hin} &
\textbf{ind} &
\textbf{jpn} &
\textbf{kor} &
\textbf{rus} &
\textbf{spa} &
\textbf{swa} &
\textbf{tel} &
\textbf{tha} &
\textbf{yor} &
\textbf{zho} &
\textbf{avg.} \\

\midrule

CLS & -- & CLS & -- 
& 66.1 & 63.4 & 47.4 & 48.1 & 49.5 & 68.4 & 51.6 & 44.5 & 48.4 & 59.2 & 61.7 & 58.0 & 49.2 & 68.2 & 75.9 & 69.1 & 58.4 & 50.1 & 57.6 \\

Mean & -- & Mean & -- 
& 65.0 & 61.4 & 45.7 & 44.5 & 50.5 & 67.3 & 48.8 & 43.6 & 48.2 & 58.3 & 61.4 & 54.7 & 49.3 & 68.8 & 77.3 & 66.8 & 50.1 & 47.2 & 56.1 \\

Latent Attn & -- & Latent Attn & -- 
& 65.5 & 62.9 & 46.2 & 45.6 & 50.6 & 67.7 & 49.1 & 43.2 & 48.8 & 58.6 & 61.0 & 56.0 & 50.5 & 68.9 & 77.4 & 67.5 & 51.2 & 48.7 & 56.6 \\

\midrule

LMK & 128 & LMK & 128
& 67.2 & 65.6 & 45.5 & 45.2 & 51.0 & 68.9 & 50.8 & 44.9 & 50.3 & 57.5 & 61.6 & 56.3 & 50.2 & 71.0 & 78.1 & 68.9 & 59.5 & 52.1 & 58.0 \\

\midrule

\multirow{4}{*}{LMK} &
\multirow{4}{*}{Variable} &
\multirow{4}{*}{LMK} & 32
& 66.5 & 63.7 & 45.8 & 44.7 & 50.5 & 69.1 & 50.0 & 44.8 & 50.1 & 59.9 & 60.9 & 57.6 & 49.8 & 69.9 & 77.1 & 68.9 & 61.2 & 49.8 & 57.8 \\

& &  & 64
& 66.4 & 63.3 & 46.2 & 45.3 & 50.7 & 69.0 & 49.8 & 44.9 & 50.1 & 60.4 & 61.3 & 57.7 & 49.7 & 69.8 & 77.1 & 69.1 & 61.7 & 50.2 & 57.9 \\

& &  & 128
& 66.5 & 63.8 & 46.1 & 44.8 & 50.9 & 68.9 & 50.0 & 45.2 & 49.9 & 59.8 & 61.1 & 57.4 & 50.1 & 70.1 & 77.2 & 68.9 & 61.5 & 50.9 & 57.9 \\

& & & 256
& 66.5 & 62.8 & 46.0 & 44.8 & 50.9 & 68.5 & 49.4 & 44.5 & 49.9 & 59.8 & 61.0 & 57.2 & 50.0 & 70.0 & 77.3 & 68.8 & 60.5 & 51.0 & 57.7 \\

\bottomrule
\end{tabular}
}

\caption{NDCG@10  retrieval performance on MIRACL-HN (18) dataset across languages under different pooling and granularity settings for \emph{mmBERT-base}.}
\label{tab:mmbert_miracl_results}
\end{table*}

\begin{table*}[t]
\centering
\small
\setlength{\tabcolsep}{3.5pt}
\renewcommand{\arraystretch}{1.15}

\resizebox{\textwidth}{!}{%
\begin{tabular}{l c l c
cccccccccccccc}
\toprule

\textbf{FT. Pool} &
\textbf{FT. Gran.} &
\textbf{Eval. Pool} &
\textbf{Eval. Gran.} &
\textbf{ara} &
\textbf{cmn} &
\textbf{deu} &
\textbf{eng} &
\textbf{fra} &
\textbf{hin} &
\textbf{ita} &
\textbf{jpn} &
\textbf{kor} &
\textbf{por} &
\textbf{rus} &
\textbf{spa} &
\textbf{tha} &
\textbf{avg.} \\

\midrule

CLS & -- & CLS & --
& 21.9 & 9.0 & 13.6 & 8.5 & 42.1 & 13.1 & 35.7 & 28.5 & 13.3 & 50.9 & 35.3 & 45.0 & 7.2 & 24.9 \\

Mean & -- & Mean & --
& 11.9 & 7.4 & 14.3 & 6.0 & 7.9 & 11.2 & 15.3 & 30.2 & 5.4 & 36.3 & 26.5 & 39.8 & 9.4 & 17.0 \\

Latent Attn & -- & Latent Attn & --
& 15.5 & 6.0 & 18.0 & 4.3 & 11.8 & 11.3 & 20.9 & 30.8 & 6.3 & 41.5 & 30.9 & 46.2 & 11.1 & 19.6 \\

\midrule

LMK & 128 & LMK & 128
& 31.0 & 15.5 & 31.5 & 24.7 & 60.9 & 19.3 & 48.3 & 36.8 & 21.2 & 63.4 & 46.4 & 60.6 & 23.5 & 37.2 \\

\midrule

\multirow{4}{*}{LMK} &
\multirow{4}{*}{Variable} &
\multirow{4}{*}{LMK} & 32
& 30.8 & 16.9 & 33.2 & 27.1 & 60.3 & 20.2 & 51.8 & 39.3 & 25.2 & 63.7 & 48.2 & 63.1 & 23.8 & 38.7 \\

& &  & 64
& 31.0 & 16.3 & 32.8 & 27.0 & 60.4 & 19.5 & 51.8 & 38.6 & 25.4 & 63.9 & 47.9 & 63.6 & 23.6 & 38.6 \\

& &  & 128
& 30.4 & 15.6 & 32.3 & 27.2 & 60.9 & 18.5 & 52.1 & 39.5 & 24.0 & 63.1 & 47.7 & 63.4 & 23.7 & 38.3 \\

& &  & 256
& 30.2 & 14.3 & 32.4 & 27.4 & 60.7 & 18.2 & 51.8 & 39.8 & 23.1 & 64.0 & 46.2 & 63.6 & 21.8 & 38.0 \\

\bottomrule
\end{tabular}
}

\caption{NDCG@10  retrieval performance on MLDR (13) dataset across languages under different pooling and granularity settings for \emph{mmBERT-base}.}
\label{tab:mmbert_mldr_results}
\end{table*}

\begin{table*}[ht]
\centering
\small
\setlength{\tabcolsep}{3.5pt}
\renewcommand{\arraystretch}{1.15}

\resizebox{\textwidth}{!}{%
\begin{tabular}{l c l c
cccccccccccccccccccccccc}
\toprule

\textbf{FT. Pool} &
\textbf{FT. Gran.} &
\textbf{Eval. Pool} &
\textbf{Eval. Gran.} &
\textbf{deu} &
\textbf{eng} &
\textbf{bul} &
\textbf{por} &
\textbf{lit} &
\textbf{swe} &
\textbf{est} &
\textbf{spa} &
\textbf{nld} &
\textbf{pol} &
\textbf{fra} &
\textbf{slk} &
\textbf{fin} &
\textbf{mlt} &
\textbf{ita} &
\textbf{hun} &
\textbf{lav} &
\textbf{ces} &
\textbf{ell} &
\textbf{slv} &
\textbf{dan} &
\textbf{ron} &
\textbf{hrv} &
\textbf{avg.} \\

\midrule

CLS & -- & CLS & --
& 27.2 & 26.8 & 28.7 & 27.3 & 29.6 & 26.8 & 27.6 & 26.0 & 26.4 & 28.5 & 27.7 & 27.3 & 27.1 & 24.6 & 27.3 & 29.1 & 30.5 & 30.4 & 25.1 & 26.9 & 27.6 & 29.3 & 25.2 & 27.5 \\

Mean & -- & Mean & --
& 27.3 & 27.9 & 30.3 & 27.0 & 29.0 & 26.9 & 28.7 & 26.9 & 25.8 & 28.6 & 26.5 & 27.7 & 26.5 & 24.9 & 26.7 & 28.8 & 30.6 & 29.9 & 25.9 & 27.4 & 26.3 & 29.2 & 23.5 & 27.5 \\

Latent Attn & -- & Latent Attn & --
& 28.1 & 27.3 & 30.8 & 27.4 & 29.3 & 27.2 & 29.2 & 27.6 & 27.4 & 28.5 & 26.9 & 28.2 & 27.3 & 25.1 & 27.2 & 30.0 & 31.8 & 30.3 & 26.4 & 28.3 & 26.7 & 29.2 & 24.0 & 28.0 \\

\midrule

LMK & 128 & LMK & 128
& 31.5 & 30.0 & 33.5 & 29.4 & 33.1 & 31.3 & 30.3 & 30.5 & 30.3 & 33.1 & 31.9 & 31.8 & 31.3 & 29.5 & 30.3 & 33.4 & 33.9 & 34.2 & 28.5 & 31.6 & 31.2 & 33.6 & 29.8 & 31.5 \\

\midrule

\multirow{4}{*}{LMK} &
\multirow{4}{*}{Variable} &
\multirow{4}{*}{LMK} & 32
& 34.1 & 32.1 & 34.9 & 31.6 & 34.8 & 32.9 & 32.8 & 32.3 & 33.2 & 35.5 & 34.0 & 33.8 & 33.7 & 32.5 & 32.5 & 35.4 & 35.4 & 36.7 & 30.3 & 34.2 & 33.4 & 35.2 & 32.6 & 33.7 \\

& & & 64
& 34.0 & 31.8 & 34.9 & 31.3 & 34.7 & 32.6 & 32.6 & 32.1 & 32.8 & 35.3 & 33.8 & 33.8 & 33.1 & 32.2 & 32.3 & 35.1 & 35.4 & 36.4 & 30.3 & 33.9 & 33.0 & 35.0 & 32.1 & 33.4 \\

& & & 128
& 34.1 & 31.6 & 34.8 & 31.2 & 34.7 & 32.5 & 32.4 & 31.8 & 32.6 & 35.2 & 33.8 & 33.7 & 32.8 & 31.9 & 32.4 & 34.8 & 35.6 & 36.0 & 30.2 & 33.7 & 32.7 & 34.8 & 31.3 & 33.2 \\

& & & 256
& 34.0 & 31.5 & 34.5 & 30.9 & 34.4 & 32.3 & 32.4 & 31.5 & 32.4 & 35.1 & 33.4 & 33.4 & 32.5 & 31.4 & 32.3 & 34.5 & 35.3 & 35.7 & 29.9 & 33.2 & 32.3 & 34.6 & 30.9 & 33.0 \\

\bottomrule
\end{tabular}
}
\caption{Zero-shot long-document classification F1 scores on MultiEURLEX across 23 languages using fine-tuned \emph{mmBERT-base}, evaluated with a maximum sequence length of 8192.}
\label{tab:mmbert_multieurlex_results}
\end{table*}

\begin{table*}[ht]
\centering
\small
\setlength{\tabcolsep}{5pt}
\renewcommand{\arraystretch}{1.15}

\resizebox{\textwidth}{!}{%
\begin{tabular}{l c l c l c
                c c
                c c c c
                c c c c}
\toprule

 & & & & & &
\multicolumn{2}{c}{\textsc{In-Domain}} &
\multicolumn{4}{c}{\textsc{Out-of-Domain (Short Context)}} &
\multicolumn{4}{c}{\textsc{Out-of-Domain (Long Context)}} \\

\cmidrule(lr){7-8}
\cmidrule(lr){9-12}
\cmidrule(lr){13-16}

 & & & & & &
\multicolumn{2}{c}{\textbf{MSMarco}} &
\textbf{BEIR} &
\textbf{MTEBv2} &
\textbf{MIRACL-HN} &
\textbf{Avg.} &
\textbf{MLDR} &
\textbf{COIR} &
\textbf{LongEmbed} &
\textbf{Avg.} \\

 \multicolumn{2}{c}{\textsc{Pretraining}} &
\multicolumn{2}{c}{\textsc{Finetuning}} &
\multicolumn{2}{c}{\textsc{Evaluation}} &
\multicolumn{2}{c}{\textbf{Dev}} &
\textbf{(15)} &
\textbf{(10)} &
\textbf{EN} &
\textbf{Short} &
\textbf{EN} &
\textbf{(8)} &
\textbf{(6)} &
\textbf{Long} \\

\cmidrule(lr){1-2}
\cmidrule(lr){3-4}
\cmidrule(lr){5-6}

\textbf{Pool} &
\textbf{Gran.} &
\textbf{Pool} &
\textbf{Gran.} &
\textbf{Pool} &
\textbf{Gran.} &
\texttt{NDCG@10} &
\texttt{MRR@10} &
\texttt{NDCG@10} &
\texttt{NDCG@10} &
\texttt{NDCG@10} &
 &
\texttt{NDCG@10} &
\texttt{NDCG@10} &
\texttt{P@1\;|\;NDCG@10} &
 \\

\midrule

CLS & -- & CLS & -- & CLS & --
& 42.5 & 36.1 & 44.7 & 45.5 & 47.4 & 45.9
& 31.2 & 42.1 & 60.8 & 44.7 \\

\midrule

\multirow{3}{*}{LMK}  & 64 & \multirow{3}{*}{LMK} & 64 & \multirow{3}{*}{LMK} & 64
& 42.6 & 36.2 & 44.5 & 45.5 & 46.8 & 45.6
& 36.7 & 43.0 & 66.1 & 48.6 \\

 & 128 &  & 128 &  & 128
& 42.8 & 36.5 & 44.2 & 44.8 & 47.3 & 45.4
& 34.6 & 43.2 & 65.7 & 47.8 \\

 & 256 &  & 256 &  & 256
& 42.8 & 36.3 & 44.5 & 45.1 & 48.7 & 46.1
& 36.0 & 42.8 & 67.6 & 48.8 \\

\midrule

\multirow{4}{*}{LMK} &
\multirow{4}{*}{Variable} &
\multirow{4}{*}{LMK} & \multirow{4}{*}{Variable} & \multirow{4}{*}{LMK} & 32
& 42.6 & 36.1 & 44.3 & 45.4 & 46.7 & 45.5
& 37.8 & 43.5 & 68.4 & 49.9 \\

& & & & & 64
& 42.5 & 36.2 & 44.3 & 45.3 & 46.5 & 45.4
& 38.1 & 43.3 & 68.4 & 49.9 \\

& & & & & 128
& 42.7 & 36.3 & 44.4 & 45.4 & 47.1 & 45.6
& 37.5 & 43.2 & 68.1 & 49.6 \\

& & & & & 256
& 42.6 & 36.2 & 44.5 & 45.4 & 48.4 & 46.1
& 36.6 & 43.3 & 67.7 & 49.2 \\

\midrule

LMK & Sentence & LMK & Sentence & LMK & Sentence
& 42.5 & 35.9 & 44.5 & 45.3 & 47.4 & 45.7
& 36.2 & 42.2 & 63.6 & 47.3 \\

\bottomrule
\end{tabular}
}
\caption{Comparison of pooling strategies on in-domain and out-of-domain retrieval benchmarks for \emph{gte-en-mlm-base}, pretrained with \texttt{CLS} or \lmk pooling and fine-tuned on MS MARCO passage data.}
\label{tab:gte_en_pretrain_results}
\end{table*}

\begin{table*}[ht]
\centering
\small
\setlength{\tabcolsep}{5pt}
\renewcommand{\arraystretch}{1.15}

\resizebox{\textwidth}{!}{%
\begin{tabular}{l c l c l c
                c
                c c c c
                c c c c}
\toprule

 & & & & & &
\multicolumn{1}{c}{\textsc{In-Domain}} &
\multicolumn{4}{c}{\textsc{Out-of-Domain (Short Context)}} &
\multicolumn{4}{c}{\textsc{Out-of-Domain (Long Context)}} \\

\cmidrule(lr){7-7}
\cmidrule(lr){8-11}
\cmidrule(lr){12-15}

 & & & & & &
\textbf{MSMarco} &
\textbf{BEIR} &
\textbf{MTEBv2} &
\textbf{MIRACL-HN} &
\textbf{Avg.} &
\textbf{MLDR} &
\textbf{COIR} &
\textbf{LongEmbed} &
\textbf{Avg.} \\

 \multicolumn{2}{c}{\textsc{Pretraining}} &
\multicolumn{2}{c}{\textsc{Finetuning}} &
\multicolumn{2}{c}{\textsc{Evaluation}} &
\textbf{Dev} &
\textbf{(15)} &
\textbf{(10)} &
\textbf{EN} &
\textbf{Short} &
\textbf{EN} &
\textbf{(8)} &
\textbf{(6)} &
\textbf{Long} \\

\cmidrule(lr){1-2}
\cmidrule(lr){3-4}
\cmidrule(lr){5-6}

\textbf{Pool} &
\textbf{Gran.} &
\textbf{Pool} &
\textbf{Gran.} &
\textbf{Pool} &
\textbf{Gran.} &
\texttt{NDCG@10} &
\texttt{NDCG@10} &
\texttt{NDCG@10} &
\texttt{NDCG@10} &
 &
\texttt{NDCG@10} &
\texttt{NDCG@10} &
\texttt{P@1\;|\;NDCG@10} &
 \\

\midrule

CLS & -- & CLS & -- & CLS & --
& 39.8 & 42.8 & 44.4 & 47.9 & 45.0
& 24.9 & 44.3 & 42.1 & 37.1 \\

\midrule

LMK & 64 & LMK & 32 & LMK & 32
& 40.3 & 42.5 & 44.1 & 47.7 & 44.8
& 31.9 & 46.8 & 59.8 & 46.2 \\

LMK & Variable & LMK & Variable & LMK & 32
& 40.2 & 41.3 & 42.5 & 47.8 & 43.9
& 31.8 & 46.2 & 66.4 & 48.1 \\

LMK & Sentence & LMK & Sentence & LMK & Sentence
& 40.0 & 42.0 & 43.2 & 46.9 & 44.0
& 32.3 & 43.4 & 64.9 & 46.9 \\

\bottomrule
\end{tabular}
}

\caption{Comparison of pooling strategies on in-domain and out-of-domain retrieval benchmarks for \emph{ModernBERT-base}, pretrained with \texttt{CLS} or \lmk pooling and fine-tuned on MS MARCO passage data.}
\label{tab:modernbert_en_pretrain_results}
\end{table*}

\begin{table*}[ht]
\centering
\small
\setlength{\tabcolsep}{5pt}
\renewcommand{\arraystretch}{1.15}

\resizebox{0.8\linewidth}{!}{%
\begin{tabular}{c cc cc c c c c c c c c}
\toprule

\multirow{2}{*}{\textbf{Model}} &
\multicolumn{2}{c}{\textsc{Finetune}} &
\multicolumn{2}{c}{\textsc{Eval}} &
\textbf{Class.} &
\textbf{Clust.} &
\textbf{PairCls.} &
\textbf{Rerank.} &
\textbf{Retr.} &
\textbf{STS} &
\textbf{Summ.} &
\textbf{Avg.} \\

\cmidrule(lr){2-3}
\cmidrule(lr){4-5}

& \textbf{Pool} & \textbf{Gran.}
& \textbf{Pool} & \textbf{Gran.}
& \textbf{(8)} & \textbf{(8)} & \textbf{(3)} & \textbf{(2)}
& \textbf{(10)} & \textbf{(9)} & \textbf{(1)} & \textbf{(41)} \\
\midrule

\multirow{3}{*}{\emph{gte-en-mlm-base}}
& CLS  & --  & CLS  & --  & 69.5 & 41.6 & 83.3 & 44.8 & 44.7 & 77.1 & 27.6 & 58.5 \\
& Mean & --  & Mean & --  & 70.1 & 40.9 & 83.3 & 44.5 & 44.9 & 75.9 & 30.4 & 58.3 \\
& LMK  & 256 & LMK  & 256 & 70.0 & 41.5 & 83.3 & 44.9 & 44.6 & 76.5 & 27.2 & 58.4 \\
\midrule

\multirow{4}{*}{\emph{ModernBERT-base}}
& CLS  & --       & CLS  & --       & 69.5 & 41.6 & 80.4 & 44.1 & 44.2 & 76.0 & 24.2 & 57.8 \\
& Mean & --       & Mean & --       & 69.6 & 41.2 & 79.4 & 43.6 & 43.3 & 75.3 & 24.8 & 57.3 \\
& LMK  & 32       & LMK  & 32       & 69.9 & 42.2 & 80.2 & 44.3 & 45.9 & 74.9 & 25.6 & 58.1 \\
& LMK  & Variable & LMK  & 128      & 69.9 & 42.5 & 80.5 & 44.1 & 45.1 & 75.2 & 25.7 & 58.1 \\

\bottomrule
\end{tabular}
}
\caption{MTEB-v2 embedding benchmark results. We report task-wise averages across pooling strategies for \emph{gte-en-mlm-base} and \emph{ModernBERT-base}, trained on MSMarco Passage data.}
\label{tab:mtebv2_gte_modernbert_msmarco_psg}
\end{table*}

\section{Training and Evaluation Hyperparameters}
\label{app:train_eval_details}

Our experiments are divided into English and Multilingual settings, with both short- and long-context training and evaluation.

\paragraph{English:} 
We use Transformer-based text encoders supporting long-context training, namely \emph{\href{https://huggingface.co/Alibaba-NLP/gte-en-mlm-base}{gte-en-mlm-base}} and \emph{\href{https://huggingface.co/answerdotai/ModernBERT-base}{ModernBERT-base}}. Models are trained on MS MARCO passage and document ranking datasets using hard negatives (7 for passage, 1 for document) and distillation scores provided by \citet{bge-en-icl-paper}. Training is performed for 5k steps with a learning rate of $2\times10^{-5}$, warmup steps of 250, and query maximum length capped at 128 tokens. Passage ranking uses an effective query batch size of 2,048 and maximum sequence length of 512 tokens; document ranking uses batch size 256 and sequences up to 8,192 tokens.  

\paragraph{Multilingual:} 
We use \emph{\href{https://huggingface.co/jhu-clsp/mmBERT-base}{mmBERT-base}} and \emph{\href{https://huggingface.co/Alibaba-NLP/gte-multilingual-base}{gte-multilingual-base}}. Models are fine-tuned on multilingual datasets from \citet{bge-m3} with hard negatives (7) and distillation scores. Training is performed for 10k steps with learning rate $2\times10^{-5}$, batch size 1,024, and maximum sequence length 512 tokens. For ablations with sequence lengths up to 8{,}192 tokens, batch size is reduced to 64 due to memory constraints.

\paragraph{Training setup:} 
We build our training pipeline on top of FlagEmbedding\footnote{\href{https://github.com/FlagOpen/FlagEmbedding/tree/master/research/baai_general_embedding}{github.com/FlagOpen/FlagEmbedding}}. We use BF16 precision on 8 NVIDIA H100 80GB GPUs. InfoNCE loss temperature is set to $\tau=0.02$, and negatives are shared across devices to increase effective in-batch negative size. Training is optimized using DeepSpeed\footnote{\href{https://github.com/deepspeedai/DeepSpeed}{github.com/deepspeedai/DeepSpeed}} and HuggingFace Accelerate\footnote{\href{https://github.com/huggingface/accelerate}{github.com/huggingface/accelerate}}.

\paragraph{Evaluation:} 
All embedding evaluations are conducted using MTEB\footnote{\href{https://github.com/embeddings-benchmark/mteb}{github.com/embeddings-benchmark/mteb}} and SentenceTransformers\footnote{\href{https://github.com/huggingface/sentence-transformers}{huggingface/sentence-transformers}}. For English, short-context benchmarks include BEIR-15, MTEBv2, and MIRACL Retrieval; long-context benchmarks include MLDR, COIR, and LongEmbed. For multilingual experiments, short-context evaluation uses MIRACL Hard Negatives (18 languages) and long-context evaluation uses MLDR (13 languages) and LongEmbed. Zero-shot long-document classification is evaluated on Multi-EURLEX (23 languages) using Macro-F1.

\section{Dataset Information}
\label{app:dataset_details}

We begin with standard English retrieval training datasets commonly used in dense retrieval, such as MSMarco Passage Ranking, TREC-DL, and Natural Questions (NQ) \cite{msmarco_dataset,trec_dl_paper,kwiatkowski-etal-2019-natural}. These datasets have been instrumental in training effective short-context retrievers; however, they primarily consist of relatively short passages and therefore do not expose models to long document structures during training.

To partially address this limitation, the MSMarco Document Ranking dataset is often used, as it contains longer documents compared to passage-level data. However, we find that this dataset remains limited in terms of true long-context coverage, with document lengths falling well below those encountered in many real-world retrieval scenarios (\cref{fig:msmarco_doc_length_plot}).

More recently, embedding training datasets have diversified substantially. Modern training mixtures, such as those used in the \emph{bge-en-icl} framework \cite{bge-en-icl-paper}, combine heterogeneous data sources spanning multiple tasks, including semantic textual similarity, clustering, retrieval, classification, question answering, and natural language inference. Statistics for these datasets are reported in \cref{tab:dataset_statistic_bge_en_icl}, where we show the number of samples and character length distributions, including the median, 25th, and 75th percentiles. Even under a conservative estimate of four to five characters per token, the majority of these datasets have median sequence lengths of at most 250 tokens. The MSMarco Document dataset remains an exception, yet even at the 75th percentile its length is typically below 2000 tokens. This highlights that, despite the increased diversity of modern embedding training corpora, long-context supervision remains scarce in English retrieval training data.

Training directly on long-context inputs is further constrained by practical considerations, including increased computational cost, reduced batch sizes, and longer training times. \textbf{\lmk pooling mitigates these limitations by enabling strong long-context extrapolation even when trained predominantly on short-context data}, as demonstrated consistently across our experimental results.

For evaluation, we first report in-domain performance on the MSMarco Dev set and out-of-domain performance on the BEIR-15 benchmark, which is widely used for assessing retrieval generalization. We additionally evaluate on the MTEB-v2 retrieval subset, consisting of ten retrieval datasets designed to enable faster yet representative benchmarking \cite{mmteb}, as well as the MIRACL English Dev set, which is constructed from human annotated relevance judgments over Wikipedia articles. Since these benchmarks primarily consist of short documents (most of which have of at most 512 tokens), they are insufficient for evaluating long-context retrieval behavior. To address this limitation, we further include the MLDR English Test set, the COIR code retrieval benchmark spanning diverse programming tasks and languages, and the LongEmbed benchmark, which focuses on extremely long documents with an average length exceeding 5.5k words. Together, these datasets enable a comprehensive evaluation of retrieval performance under long-context settings, including sequence lengths beyond 32k tokens.

For multilingual training, we use the \emph{BGE-m3} training data introduced by \citet{bge-m3}, which comprises a diverse collection of datasets spanning a wide range of languages, as summarized in \cref{tab:dataset_statistic_bgem3}. This mixture includes the MLDR training set, which contains substantially longer documents than MSMarco Document Ranking and is therefore used for our long-context training experiments. However, MLDR constitutes only about 42k query document pairs out of the roughly three million total training pairs, limiting its overall influence during training and resulting in relatively weak long-context extrapolation when using conventional pooling methods.

For multilingual evaluation, we use all 18 languages available in the MIRACL Dev set for short-context tasks and the MLDR 13-language test set for long-context tasks. Since the corresponding training sets were already seen during model training, we additionally evaluate out-of-domain long-context performance. To this end, we include the Multi-EURLEX dataset, which contains 65k EU legal documents in 23 languages labeled with EUROVOC concepts by the EU Publication Office. For this out-of-domain evaluation, we report the F1 score. Finally, we include LongEmbed as a long-context benchmark to assess the models’ ability to handle long-context retrieval tasks.

\begin{table*}[t]
\centering
\small
\setlength{\tabcolsep}{6pt}
\renewcommand{\arraystretch}{1.1}

\resizebox{0.5\textwidth}{!}{%
\begin{tabular}{lrrrr}
\toprule
\textbf{Dataset} & \textbf{Count} & \textbf{Median} & \textbf{P25} & \textbf{P75} \\
\midrule
amazon counterfactual classification & 3{,}907 & 80 & 80 & 80 \\
amazon reviews classification & 20{,}000 & 78 & 67 & 78 \\
arXiv abstract & 20{,}000 & 825 & 581 & 1{,}135 \\
arguana & 4{,}065 & 813 & 595 & 1{,}098 \\
arxiv title & 20{,}000 & 66 & 50 & 86 \\
banking classification & 10{,}003 & 35 & 27 & 41 \\
biorxiv abstract & 4{,}070 & 1{,}614.5 & 1{,}325 & 1{,}975.75 \\
biorxiv title & 4{,}070 & 102 & 83 & 125 \\
eli5 & 325{,}475 & 393 & 232 & 621 \\
emotion classification & 14{,}341 & 46 & 45 & 49 \\
fever & 29{,}096 & 1{,}633 & 859 & 2{,}496 \\
fiqa & 5{,}500 & 870.5 & 503 & 1{,}453.25 \\
hotpotqa & 84{,}516 & 396 & 270 & 559 \\
imdb classification & 25{,}000 & 57 & 57 & 57 \\
medrxiv abstract & 1{,}157 & 1{,}836 & 1{,}440 & 2{,}271 \\
medrxiv title & 1{,}157 & 113 & 90 & 141 \\
msmarco document & 367{,}013 & 3{,}897 & 2{,}160 & 7{,}758 \\
msmarco passage & 485{,}823 & 336 & 281 & 454 \\
mtop intent classification & 15{,}667 & 18 & 16 & 26 \\
nli & 275{,}601 & 37 & 27 & 55 \\
nq & 58{,}568 & 624 & 599 & 652 \\
quora & 60{,}202 & 47 & 37 & 61 \\
reddit clustering & 20{,}000 & 49 & 32 & 70 \\
reddit clusteringP2P & 20{,}000 & 493 & 306 & 850.25 \\
scidocsrr & 12{,}655 & 1{,}006 & 753 & 1{,}299 \\
squad & 87{,}599 & 693 & 559 & 895 \\
stack exchange clustering & 20{,}000 & 46 & 31 & 64 \\
stack exchange clusteringP2P & 20{,}000 & 649 & 407 & 1{,}049 \\
stack overflow dup questions & 19{,}847 & 47 & 36 & 60 \\
sts & 2{,}006 & 56 & 39 & 122 \\
toxic conversations classification & 25{,}000 & 26 & 26 & 26 \\
trivial & 60{,}315 & 625 & 601 & 652 \\
tweet sentiment extraction classification & 27{,}481 & 114 & 110 & 114 \\
twenty news groups & 11{,}854 & 28 & 20 & 39 \\
\bottomrule
\end{tabular}
}

\caption{Character length statistics for positive documents in the \emph{bge-en-icl} dataset, used to train English dense retrievers by \citet{bge-en-icl-paper}. We report the number of samples and the median, 25th (P25), and 75th (P75) percentiles. Assuming $\sim$4--5 characters per token, most datasets correspond to sequence lengths well below those required for learning robust long-context representations.}
\label{tab:dataset_statistic_bge_en_icl}
\end{table*}

\begin{table*}[t]
\centering
\small
\resizebox{0.5\textwidth}{!}{%
\begin{tabular}{lrrrr}
\toprule
\textbf{Dataset} & \textbf{Count} & \textbf{Median} & \textbf{P25} & \textbf{P75} \\
\midrule
DuReader                        & 80{,}416  & 296    & 258    & 411 \\
HotpotQA                       & 84{,}516  & 396    & 270    & 559 \\
Law                            & 2{,}054   & 607    & 356.5  & 1455.75 \\
MIRACL (ar)                    & 3{,}495   & 472    & 302    & 713.5 \\
MIRACL (bn)                    & 1{,}631   & 581    & 405.5  & 831 \\
MIRACL (en)                    & 2{,}863   & 661    & 446    & 901 \\
MIRACL (es)                    & 2{,}162   & 544.5  & 376    & 758 \\
MIRACL (fa)                    & 2{,}107   & 408    & 256    & 624 \\
MIRACL (fi)                    & 2{,}897   & 466    & 332    & 651 \\
MIRACL (fr)                    & 1{,}143   & 509    & 320    & 769.5 \\
MIRACL (hi)                    & 1{,}169   & 500    & 326    & 754 \\
MIRACL (id)                    & 4{,}071   & 607    & 413    & 836 \\
MIRACL (ja)                    & 3{,}477   & 227    & 153    & 324 \\
MIRACL (ko)                    & 868       & 228    & 156    & 332 \\
MIRACL (ru)                    & 4{,}683   & 550    & 360    & 821 \\
MIRACL (sw)                    & 1{,}901   & 278    & 180    & 428 \\
MIRACL (te)                    & 3{,}452   & 566    & 395    & 946 \\
MIRACL (th)                    & 2{,}972   & 485    & 339    & 691 \\
MIRACL (zh)                    & 1{,}312   & 168    & 114    & 246 \\
MLDR (ar)                      & 1{,}817   & 13{,}535  & 5{,}174     & 21{,}921 \\
MLDR (de)                      & 1{,}847   & 24{,}850  & 15{,}070    & 29{,}238.5 \\
MLDR (en)                      & 10{,}000  & 18{,}229  & 10{,}553.75 & 25{,}636.25 \\
MLDR (es)                      & 2{,}254   & 22{,}453.5 & 6{,}579.25  & 28{,}463 \\
MLDR (fr)                      & 1{,}608   & 23{,}685  & 13{,}130.75 & 27{,}501.25 \\
MLDR (hi)                      & 1{,}618   & 12{,}323.5 & 5{,}759.75  & 20{,}120 \\
MLDR (it)                      & 2{,}151   & 25{,}551  & 12{,}454.5  & 29{,}046.5 \\
MLDR (ja)                      & 2{,}262   & 9{,}311.5 & 4{,}068.25  & 11{,}003.75 \\
MLDR (ko)                      & 2{,}198   & 7{,}529.5 & 2{,}563.25  & 11{,}906.5 \\
MLDR (pt)                      & 1{,}845   & 25{,}574  & 12{,}549    & 29{,}334 \\
MLDR (ru)                      & 1{,}864   & 18{,}851  & 7{,}516.75  & 23{,}289.5 \\
MLDR (th)                      & 1{,}970   & 10{,}283.5 & 6{,}651.25  & 22{,}316.5 \\
MLDR (zh)                      & 10{,}000  & 6{,}079.5 & 3{,}610     & 8{,}624.75 \\
MSMARCO                       & 485{,}905 & 336    & 281    & 454 \\
Mr.\ TyDi                      & 48{,}729  & 434    & 277    & 647 \\
NQ                             & 58{,}568  & 624    & 599    & 652 \\
SQuAD                          & 87{,}599  & 693    & 559    & 895 \\
T2Ranking                      & 90{,}467  & 316    & 180    & 604 \\
Trivia                         & 60{,}315  & 625    & 601    & 652 \\
cMedQAv2                       & 50{,}000  & 90     & 62     & 125 \\
EN-NLI                         & 274{,}951 & 37     & 27     & 55 \\
mMARCO                         & 100{,}000 & 110    & 84     & 152 \\
ZH-NLI (ATEC)                  & 11{,}325  & 12     & 10     & 15 \\
ZH-NLI (BQ)                    & 12{,}599  & 10     & 7      & 14 \\
ZH-NLI (LCQMC)                 & 10{,}000  & 10     & 8      & 12 \\
ZH-NLI (PAWSX)                 & 10{,}000  & 42     & 32     & 55 \\
ZH-NLI (QBQTC v2)              & 10{,}000  & 22     & 14     & 31 \\
ZH-NLI (STS)                   & 249       & 15     & 11     & 19 \\
ZH-NLI (AFQMC)                 & 10{,}534  & 12     & 10     & 16 \\
\bottomrule
\end{tabular}
}
\caption{Character length statistics for positive documents across \emph{bge-m3} dataset, used by \citet{bge-m3}. We report the number of samples along with the median, 25th (P25), and 75th (P75) percentiles. MLDR datasets feature substantially longer sequences than standard retrieval benchmarks, and without them, models using CLS or mean pooling struggle to generalize to long-context settings.}
\label{tab:dataset_statistic_bgem3}
\end{table*}

\end{document}